\documentclass{ceurart}

\copyrightyear{2024}
\copyrightclause{Copyright for this paper by its authors. Use permitted under Creative Commons License Attribution 4.0 International (CC BY 4.0).}
\conference{CLEF 2024: Conference and Labs of the Evaluation Forum, September 09–12, 2024, Grenoble, France}
\usepackage{graphicx}
\usepackage{algorithm}
\usepackage{algpseudocode}
\usepackage{amsmath}
\usepackage{placeins}
\usepackage{subfigure}
\begin{document}
\title{Tighnari: Multi-modal Plant Species Prediction Based on Hierarchical Cross-Attention Using Graph-Based and Vision Backbone-Extracted Features}
\title[mode=sub]{Notebook for the <LifeCLEF> Lab at CLEF 2024}

\author[1]{Haixu Liu}[%
orcid=0009-0007-8115-0826,
email=liuhaixu1998@foxmail.com,
]
\cormark[1]
\fnmark[1]

\author[1]{Penghao Jiang}[%
orcid=0009-0005-2930-706X,
email=pjia0498@uni.sydney.edu.au,
]
\cormark[1]
\fnmark[1]

\author[1]{Zerui Tao}[%
orcid=0009-0002-2276-6747,
email=ztao0063@uni.sydney.edu.au ,
]
\cormark[1]
\fnmark[1]

\author[1]{Muyan Wan}[%
orcid=0009-0004-5979-2943,
email=mwan3425@uni.sydney.edu.au ,
]
\cormark[1]
\fnmark[1]

\author[1]{Qiuzhuang Sun}[%
orcid=0000-0002-7103-1387,
email=qiuzhuang.sun@sydney.edu.au,
]
\cormark[1]
\fnmark[1]

\cortext[1]{Corresponding author.}
\fntext[1]{These authors contributed equally.}
\address[1]{The University of Sydney, Camperdown Campus, Sydney, 2006 NSW, Australia}

\begin{abstract}
Predicting plant species composition in specific spatiotemporal contexts plays an important role in biodiversity management and conservation, as well as in improving species identification tools. 
Our work utilizes 88,987 plant survey records conducted in specific spatiotemporal contexts across Europe.
We also use the corresponding satellite images, time series data, climate time series, and other rasterized environmental data such as land cover, human footprint, bioclimatic, and soil variables as training data to train the model to predict the outcomes of 4,716 plant surveys.
We propose a feature construction and result correction method based on the graph structure.
Through comparative experiments, we select the best-performing backbone networks for feature extraction in both temporal and image modalities. 
In this process, we built a backbone network based on the Swin-Transformer Block for extracting temporal Cubes features. 
We then design a hierarchical cross-attention mechanism capable of robustly fusing features from multiple modalities. 
During training, we adopt a 10-fold cross-fusion method based on fine-tuning and use a Threshold Top-K method for post-processing. 
Ablation experiments demonstrate the improvements in model performance brought by our proposed solution pipeline. 
This work achieves a private leaderboard score of 0.36242 in the GeoLifeCLEF 2024 \@ LifeCLEF \& CVPR-FGVC Competition,  securing third place in the rankings (Team name: Miss Qiu).
\end{abstract}


\maketitle
\section{Introduction}
\label{sec1}
\subsection{Background and related literature}
Predicting the composition of plant species over segmented time and spatial scales plays an important role in managing and protecting ecosystem biodiversity and improving species identification tools. Therefore, the LifeCLEF lab of the CLEF conference and the FGVC11 workshop of CVPR jointly host the GeoLifeCLEF 2024 competition centered around this task \cite{geolifeclef2024,lifeclef2024}.

We review the strategies of past winners in this competition. The fourth-place \cite{bib6} in 2022 mentions a method for constructing "Patched" approaches, where variables from eight different modalities provided that year are aligned into a single image format of (256,256,3). These modalities are then processed separately by ResNet50 \cite{bib10}, and the output features are concatenated and passed through a linear layer, with Top-K processing applied to the outputs. In contrast, the second-place entry from the same year \cite{bib7} abandons data from other modalities and only uses remote sensing data, creating NDVI image format features using NIR and RGB data. They train multiple models based on ResNet50, DenseNet201, and Inception-v4, and fuse the logits.The champion's strategy in 2023 \cite{bib8,lifeclef2023} introduces three backbone networks based on ResNet. The first network solely extracts features from bioclimatic raster data, while the second and third networks each include three backbones for extracting features from bioclimatic rasters, satellite images, and soil rasters, with varying depths in these two networks. Finally, logits fusion is applied to these three models. At the same time, they adopted a three-step training strategy to attempt to learn the information in PO. These successful past entries and the official baseline significantly influence our model development.

\subsection{Our method}
Our study uses survey data of 11,255 species to train a machine-learning model for this goal.
Each sample in the dataset comprises satellite imagery and time series \cite{bib1,bib2,bib3} linked with geographical coordinates, climate time series \cite{bib4,bib5}, and other rasterized environmental data such as land cover, human footprint, bioclimatic, and soil variables.

To effectively extract features across these modalities for accurate predictions, we propose the following solution pipeline for machine learning.
First, we create a graph structure based on the available data, treating SurveyIDs as nodes. 
Nodes are connected based on whether they fall within the same ecological niche (less than 10 kilometers apart) and share geographical and annual similarities. 
Subsequently, we aggregate the labels of all adjacent nodes for each node, using these as new features for that node. 
In our model, we use a Swin-Transformer-based method for extracting temporal features, instead of the commonly used Resnet18 network.
Additionally, we provide an optional replacement for the Swin-Transformer Tiny, used in the above baseline model for image feature extraction, with EfficientNet-B0. 
This substitution can significantly expedite model training with minimal impact on prediction accuracy. 
We also develop a hierarchical cross-attention mechanism (HCAM) to fuse features extracted from different modalities, which effectively uses information from multimodality data.
Last, in the post-processing steps, we improve the traditional Top-K rule for multi-task classification.
Specifically, we integrate a series of thresholds for the Top-K rule for model outputs (see Section~\ref{sec3.7}).
We then use the graph structure and the above model outputs for the final species prediction.
The details of this novel solution pipeline is provided in Section~\ref{sec3}.

\subsection{Contributions}
We summarize our contributions as follows:

\begin{itemize}
	\item [1.]	In preprocessing, we construct a graph structure for independent Survey IDs, using labels of nodes (Survey IDs) with adjacency relationships to generate new features.
    This graph structure improves the final prediction results.
	\item [2.]	Our backbone network for visual features modifies the Swin-Transformer structure to extract temporal characteristics.
    This strategy is shown to enhance model performance.
	\item [3.]	We design a hierarchical cross-attention mechanism to fuse features from different modalities.
    This strategically mitigates overfitting.
	\item [4.]	Our proposed post-processing strategy combines the advantages of the threshold method and Top-K approach, also improving the final prediction accuracies.
\end{itemize}

Our model, named Tighnari, is inspired by the Forest Ranger character in the popular open-world game ``Genshin Impact,'' who is adept at identifying a variety of species in rainforests and is committed to ecological conservation. Just as Tighnari ensures the health of the rainforest ecosystems, we aspire for our model to accurately predict the composition of plant species in given spatiotemporal contexts and make significant contributions to environmental protection. The model's acronym, TIGH, stands for T for Transformer, I for Image processing (computer vision) backbone, G for Graph feature extract, H for Hierarchical cross-attention fusion mechanism.

\section{Exploratory Data Analysis}
\label{sec2}

Exploratory Data Analysis (EDA) is essential for unveiling the motivations behind our choice of modeling techniques. We divide our analysis into several steps, starting with the visualization of unstructured data (shown in Figure \ref{fig1} and Figure \ref{fig2}):
\FloatBarrier
\begin{figure}[h]       
	\centering
	\setlength{\abovecaptionskip}{0pt}    
	\subfigure[Time series of four climatic features for four samples]{  
		\includegraphics[width=0.92\linewidth]{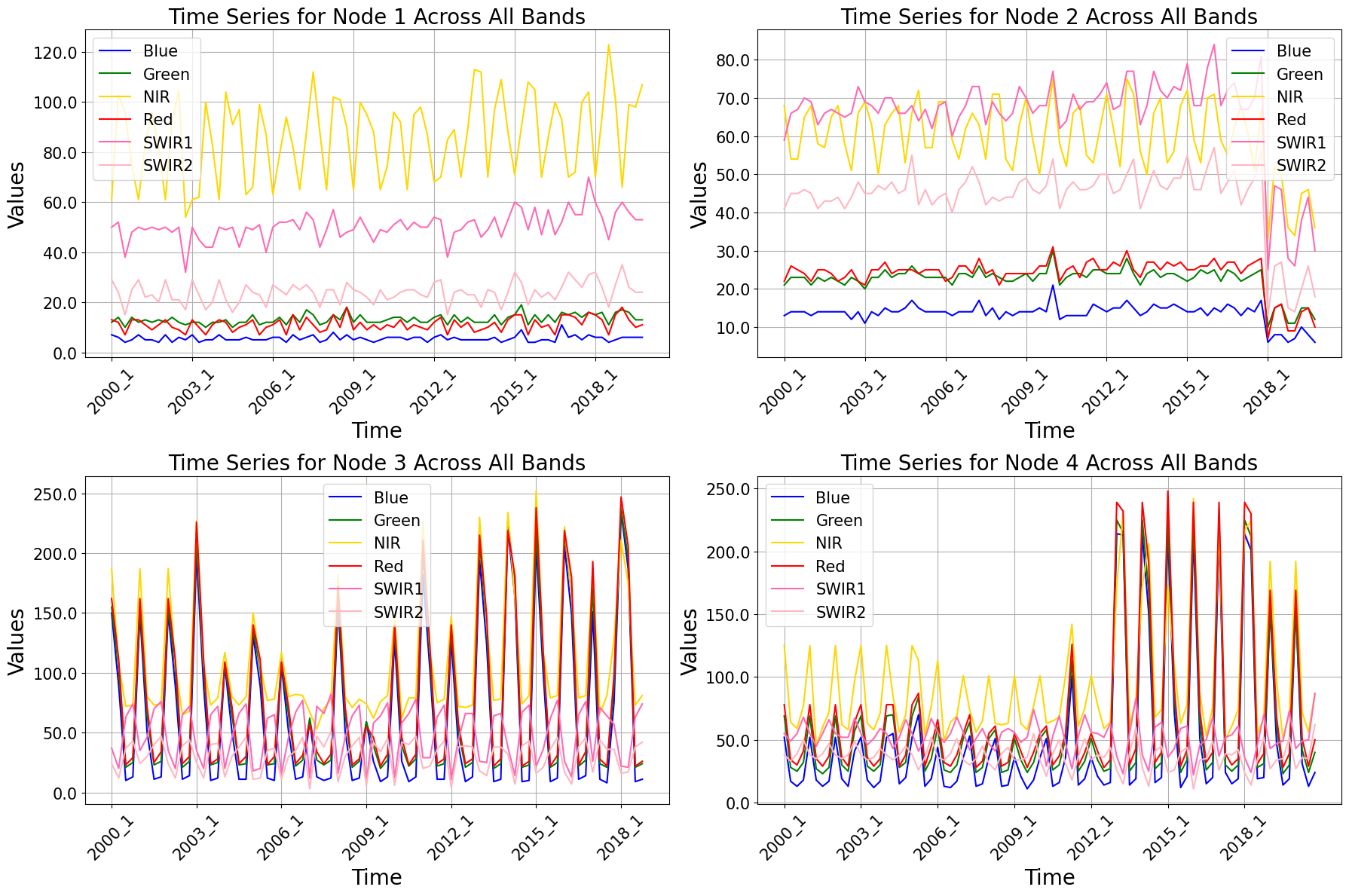}
		\label{fig11} 	}
	\subfigure[Time series of four climate features for four samples]{
		\includegraphics[width=0.92\linewidth]{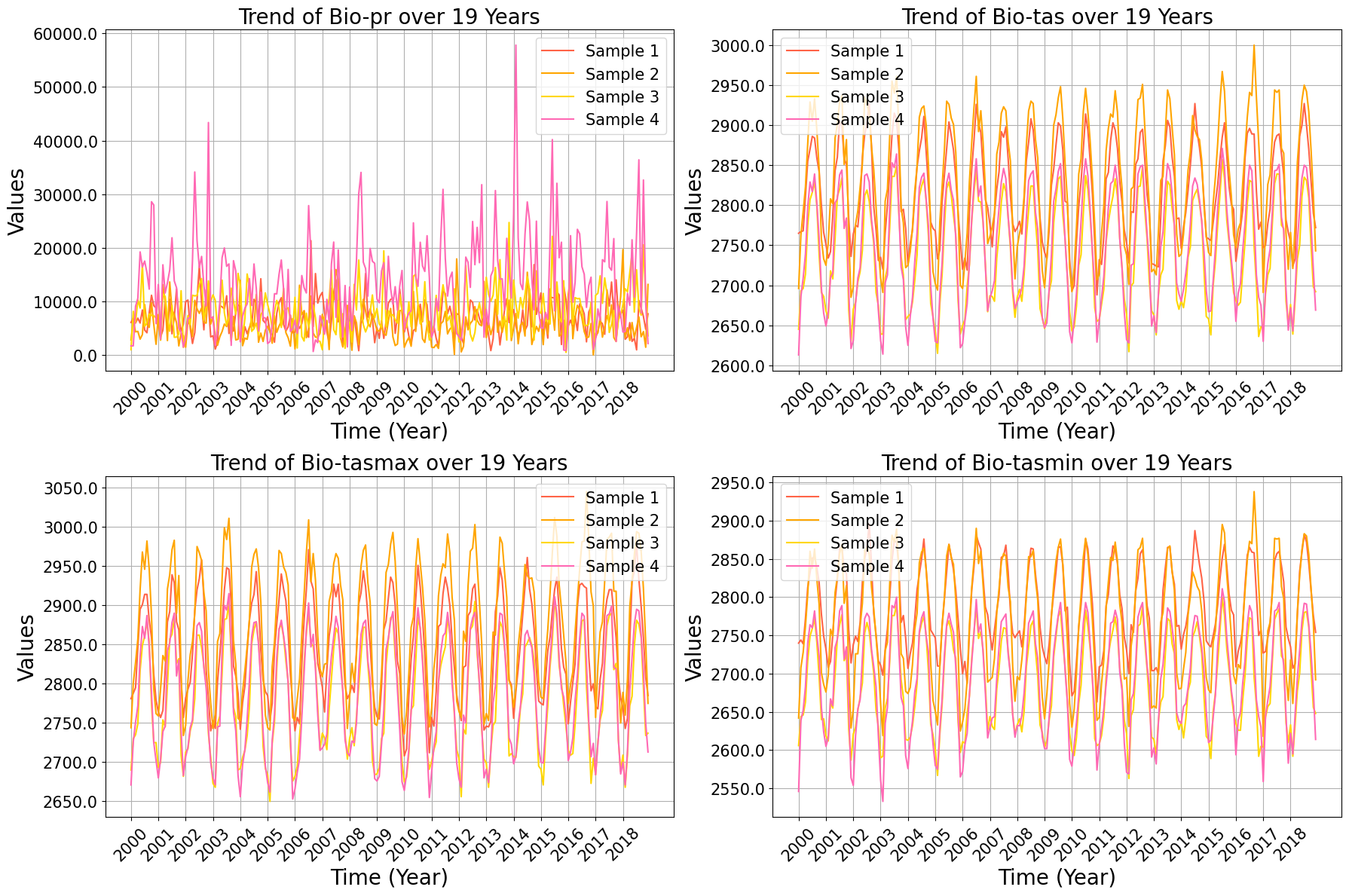}
		\label{fig12}  	}    
	\caption{Flattened visualization of time series cubes}
	\label{fig1}   
\end{figure}
\FloatBarrier
\FloatBarrier
\begin{figure}[h]     
	\centering
	\setlength{\abovecaptionskip}{0pt}    
	\subfigure[NIR image]{  
		\includegraphics[width=0.23\linewidth]{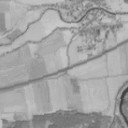}
		\label{fig21} 	}
	\subfigure[RGB image]{
		\includegraphics[width=0.23\linewidth]{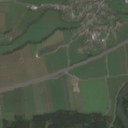}
		\label{fig22}  	}   
	\subfigure[NIR image]{
		\includegraphics[width=0.23\linewidth]{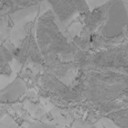}
		\label{fig23}  	}  
	\subfigure[RGB image]{
		\includegraphics[width=0.23\linewidth]{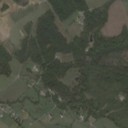}
		\label{fig24}  	}   
	\caption{Visual comparison between NIR image and RGB image}
	\label{fig2}   
\end{figure}
\FloatBarrier
We find that time series data exhibit strong periodicity, which inspires us to consider whether folding the time series according to its periodicity to transform 1D data into 2D data might facilitate more effective feature extraction.

Subsequently, we perform outlier detection in tabular data, followed by data cleaning and imputation of missing values. 
We group the data by year and region to observe the distribution of features and the correlations between them under different groupings.
\FloatBarrier
\begin{figure}[h]
	\centering
	\includegraphics[width=1\linewidth]{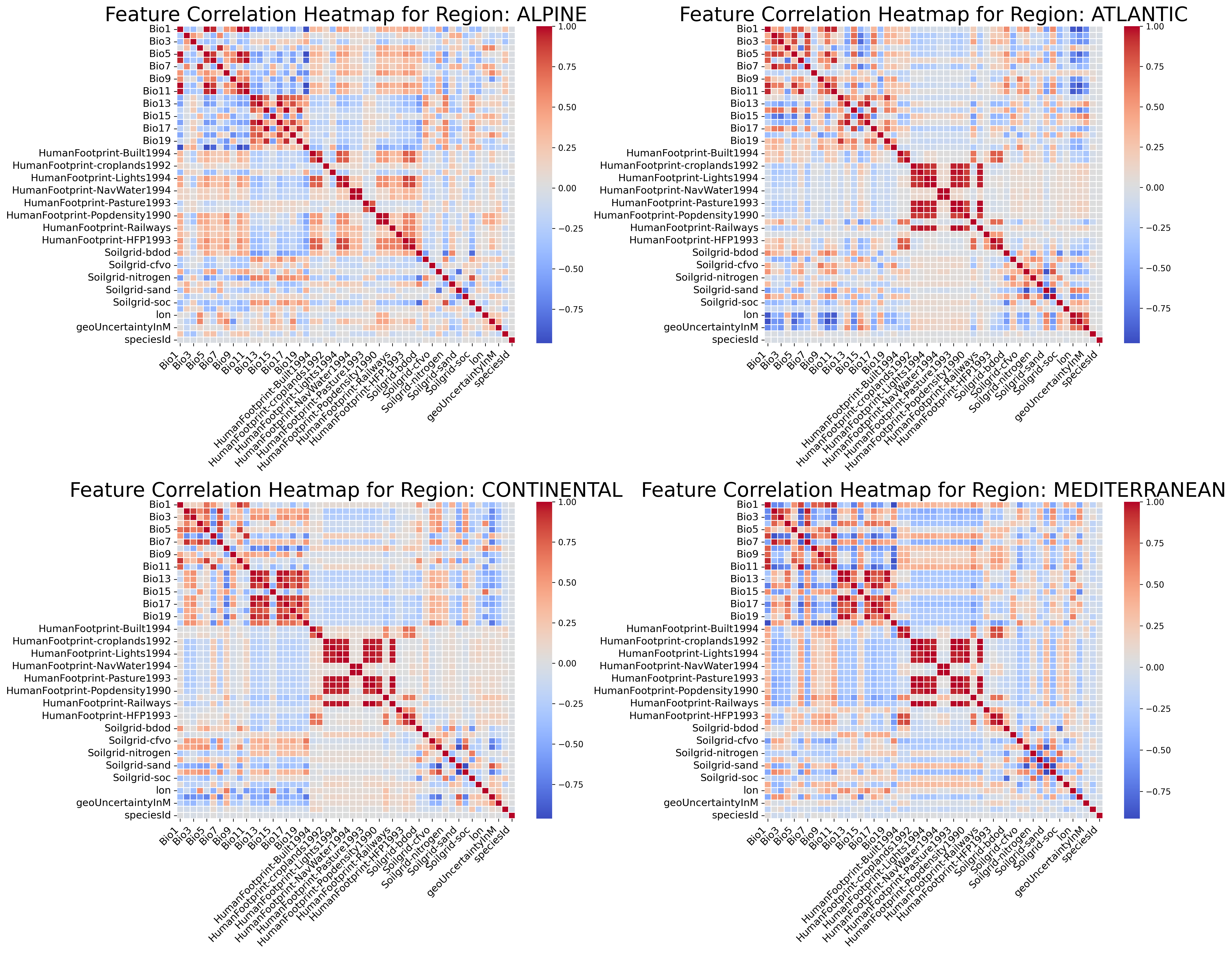}
	\caption{Comparison of numerical feature heatmaps in different regions}
	\label{fig3}
\end{figure}
\FloatBarrier
\FloatBarrier
\begin{figure}[h]
	\centering
	\includegraphics[width=1\linewidth]{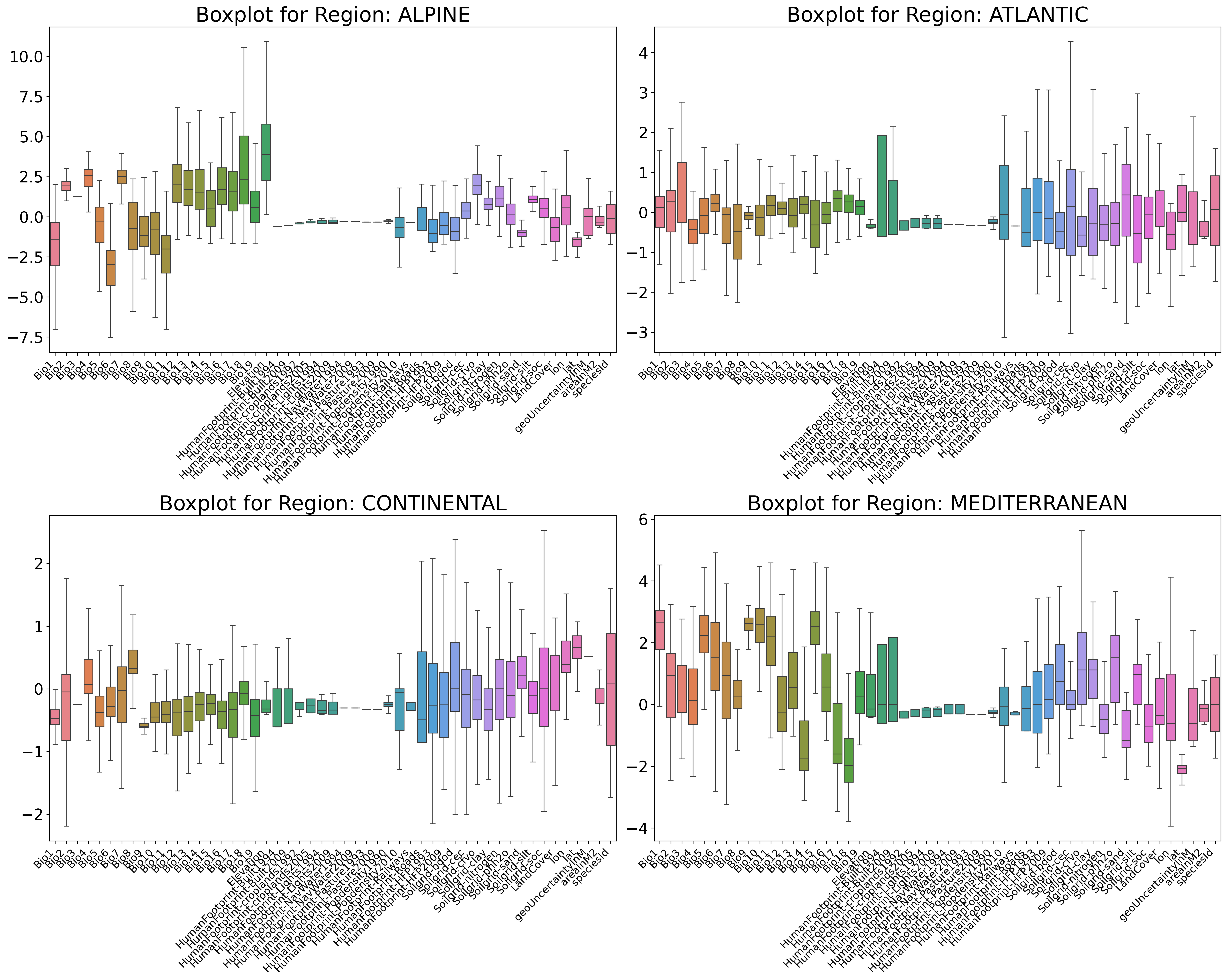}
	\caption{Comparison of numerical feature boxplots in different regions}
	\label{fig4}
\end{figure}
\FloatBarrier

The analysis reveal significant variations in feature distribution and correlations across different regions, leading us to hypothesize that these variations could influence the distribution of species (shown in Figure \ref{fig3} and Figure \ref{fig4}). This hypothesis is confirmed by further examining the prevalence of leading species in different regions. Additionally, we note that feature distribution and correlations do not show significant differences between close years, but observable differences emerge across more widely separated years. Hence, we conclude that sharing the same year and region is a prerequisite for establishing an edge between two nodes (supporting label aggregation).

\begin{figure}[h]
	\centering
	\includegraphics[width=1\linewidth]{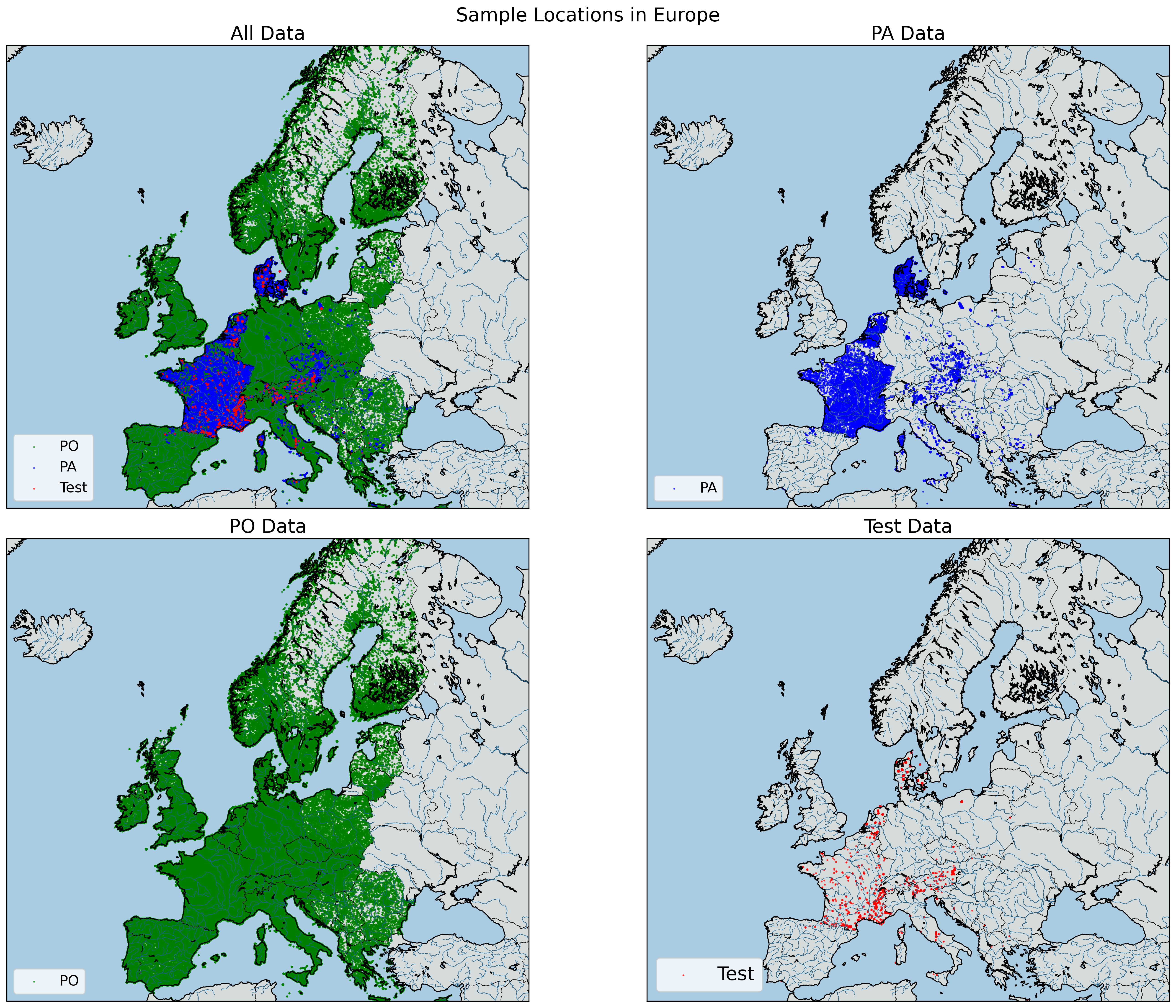}
	\caption{Visualization of survey occurrence locations in the  PO,PA as well as test set}
	\label{fig5}
\end{figure}

Next, we visualize the geographic locations of PA ad PO Survey IDs to explore whether they exhibit any clustering tendencies (shown in Figure \ref{fig5}).
\FloatBarrier
\FloatBarrier
The visualization indicates a tendency for the survey locations to cluster. We analyze some of these smaller clusters, calculating their radii, and estimated that the radius around each small cluster center was approximately 10 kilometers. To prevent a node from being overwhelmed by an excessive number of adjacent nodes, which could lead to a reduction in the variance of the aggregated feature vectors, we introduce further constraints on edge creation: no edge would exist between nodes if their geographic distance exceeded 10 kilometers.

\begin{figure}[h]       
	\centering
	\setlength{\abovecaptionskip}{0pt}
	\begin{minipage}{1\linewidth}
		\includegraphics[width=1\linewidth]{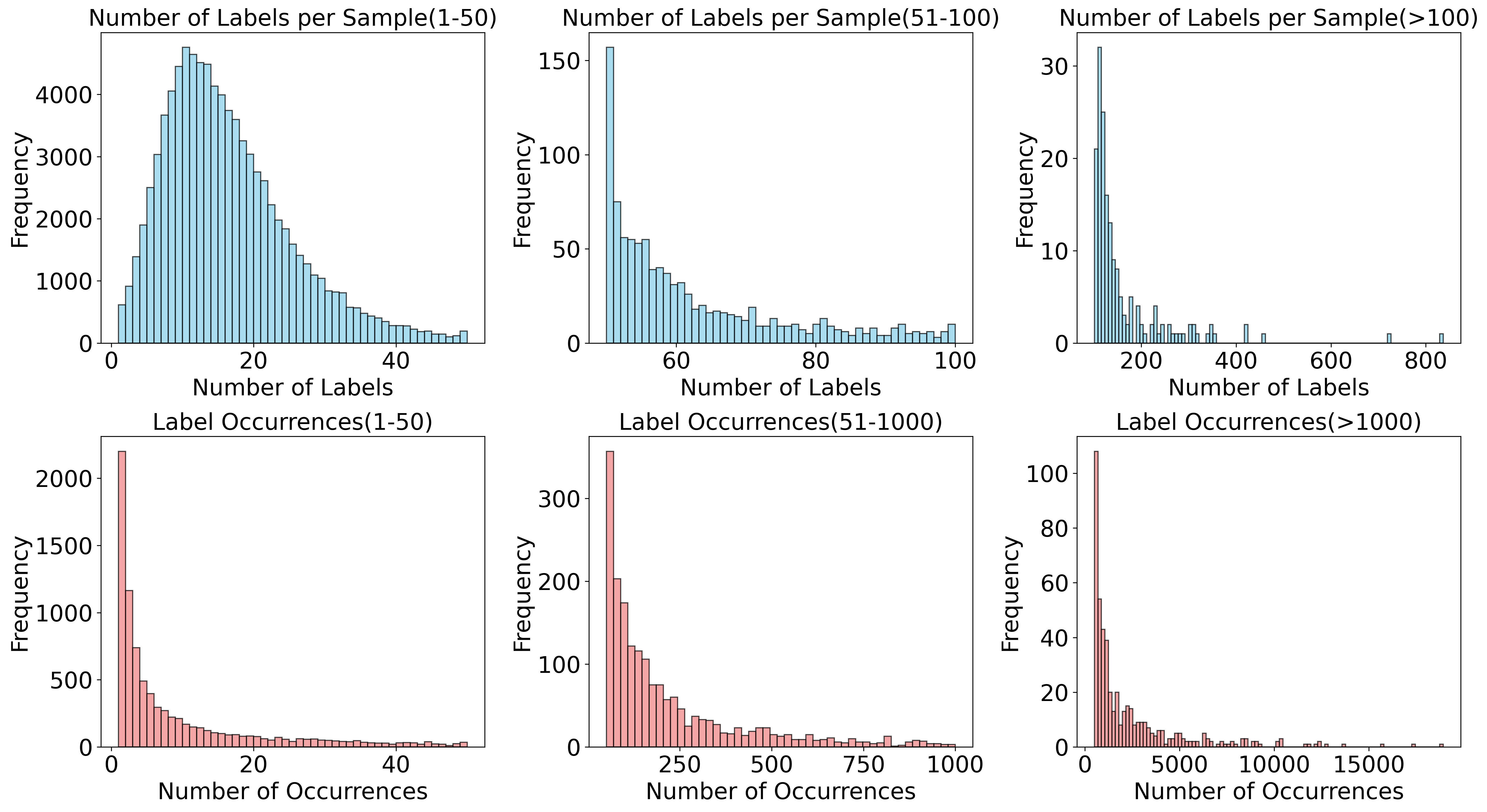}
	\end{minipage}
	\begin{minipage}{1\linewidth}
		\includegraphics[width=1\linewidth, trim=0mm 30mm 0mm 30mm, clip]{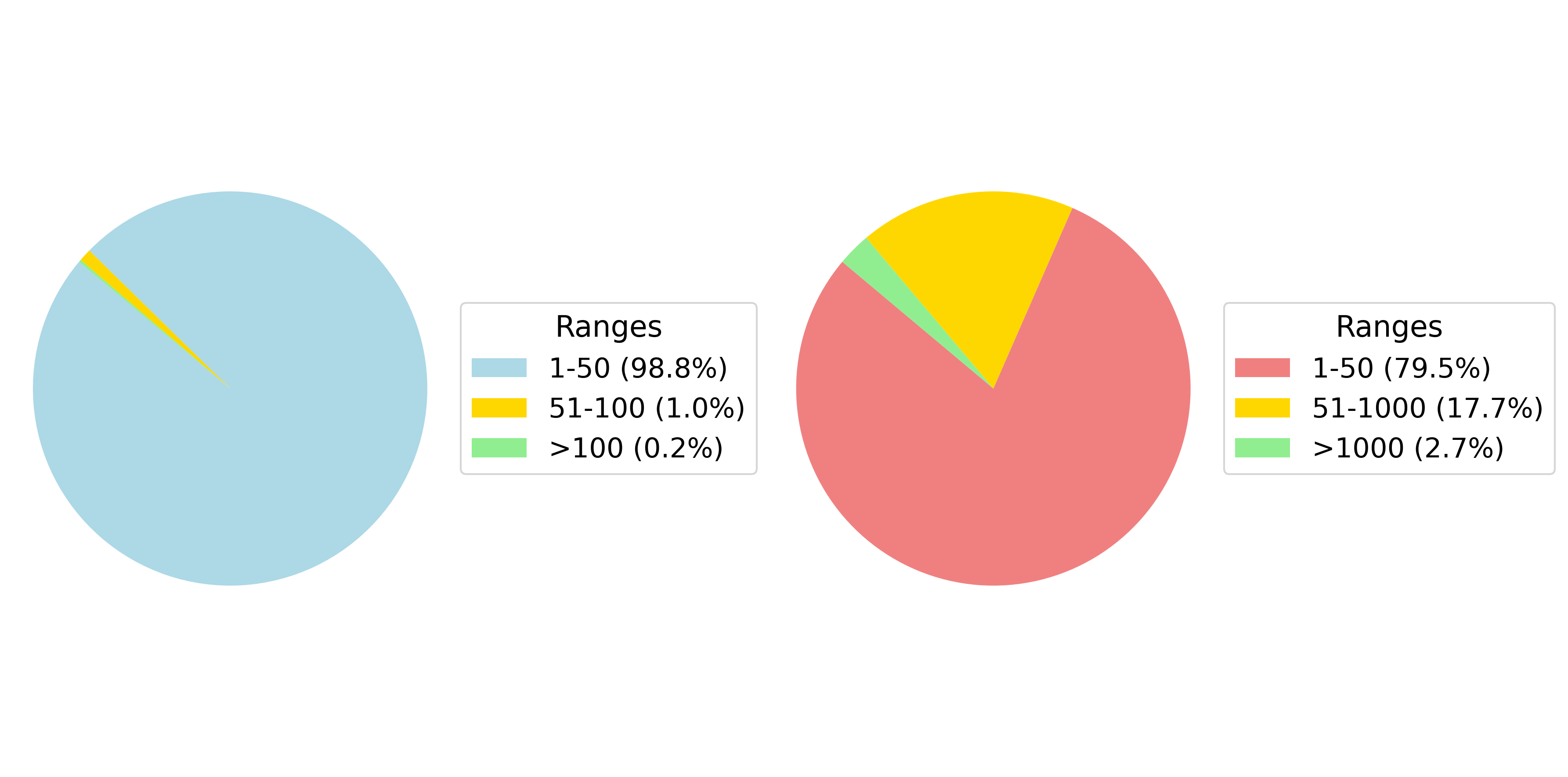}
	\end{minipage}  
	\caption{The frequency of species occurrences and the number of species included in surveys}
	\label{fig6}   
\end{figure}

We then visualize the labels, observing the frequency of species occurrences and the frequency distribution of the number of species recorded per survey (shown in Figure \ref{fig6}). This analysis aids us in setting a reasonable range for K in the Top-K process.
It can be observed that the optimal value of K should be between 0 and 100.
\FloatBarrier
\FloatBarrier

Finally, using survey IDs as nodes, we construct a graph based on the aforementioned rules. We examine the distribution of each node's degree (the sum of edge weights) on the graph. We find the distribution to be highly uneven, which could potentially degrade the quality of aggregated features on the graph. Direct normalization of the aggregated feature vectors by the degree could excessively diminish the values corresponding to rarer species. Inspired by the Attention mechanism \cite{bib18}, which normalizes attention scores using a square root transformation, we adopt this technique to use the square root of the degree as the divisor for normalizing aggregated feature vectors.
\FloatBarrier
\begin{figure}[h]
	\centering
	\includegraphics[width=1\linewidth]{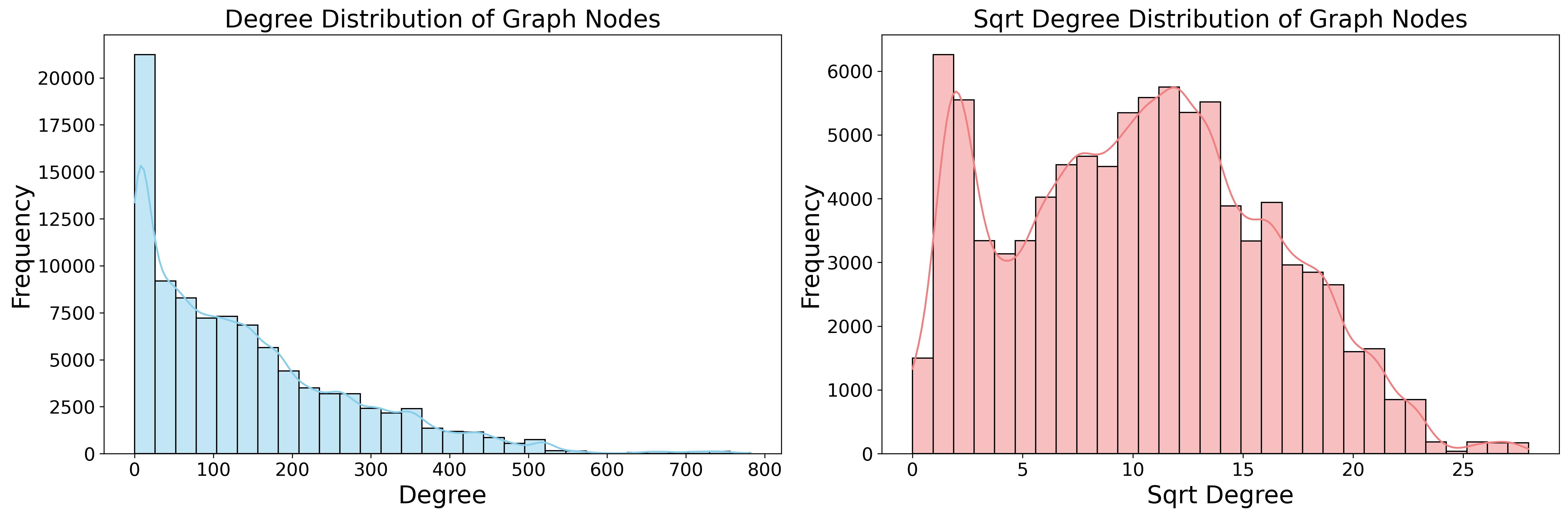}
	\caption{Distribution of node degrees in the graph and their square root transformations.}
	\label{fig7}
\end{figure}
\FloatBarrier

Through visualization, we discover that this approach allows the divisor used to normalize the feature vectors of each node in the graph to approximate a normal distribution, and also results in a smaller range of divisor (shown in Figure \ref{fig7}).

\section{Methodology}
\label{sec3}

In this section, we provide a detailed description of each step in the modeling process and the motivations behind them.

\subsection{Table Data Cleaning and Missing Value Imputation}
\label{sec3.1}
Initially, we group the metadata for both PA and PO by Survey ID. We then replace outliers in the grouped metadata of PA and the test set with null values, and impute missing values with the mean. Categorical data were then one-hot encoded. Subsequently, for the label(speciesId), we encode each element according to its corresponding number using 0-1 encoding.

Next, we access all tabular data for the training and test sets from the EnvironmentalRasters folder. We notice that the human footprint column contained $-1$ and extreme outliers. Based on the observed data distribution, we set all values greater than 255 to 255 and those less than 0 to 1. Subsequently, we merge all tables by Survey ID, replacing all infinite and infinitesimal values with nulls, and again impute all missing values with the mean. 
Finally, the training sets of the PA and test sets were merged again with the cleansed EnvironmentalRasters data by Survey ID to produce the cleaned tabular modality data.In the subsequent sections, we represent the features extracted from tables as $F$.

Following this, we clean the time series data by folding it into cubes. In the baseline method, all null values in the Tensor are replaced with zeros. We change this replaced value to the mean of the Tensor. Additionally, as the Swin-Transformer cannot accept prime numbers as the dimensions of input image matrices, we trim the shapes of two sets of time series cubes from (4, 19, 12) and (6, 4, 21) to (4, 18, 12) and (6, 4, 20) respectively. This is justified because the last year in the series already had many missing values, so trimming directly does not result in significant information loss.

\subsection{Graph Construction and Utilization}
\label{sec3.2}
Graph construction and utilization are highlights of our work. To establish graph relationships among samples, we based our approach on two fundamental assumptions. First, we assumed a clustering tendency in the spatial distribution of individual species, meaning that if a species appears in nodes surrounding a particular Survey ID, the likelihood of its occurrence at that Survey ID increases. Second, we hypothesized that within the same ecological environment, there is a correlation between the spatiotemporal distributions of different species, implying that samples (nodes) close in geographical location and year exhibit higher similarity in species composition. Our earlier data visualization efforts validated these assumptions.

Our graph construction process was two-fold. The first step involved establishing a base graph for aggregating labels from adjacent nodes. Our rationale and actions are as follows: Visualization revealed significant differences in ecological characteristics and species distribution among different regions, and even within the same region across extensive years. Consequently, we determine that nodes being in the same year and region was a prerequisite for an edge (supporting label aggregation).

However, simply adding edges between nodes of the same year and region can result in highly consistent feature vectors due to label aggregation (with minimal variance), and the varying numbers of nodes across different regions and years can lead to significant imbalances in feature vector values, thus affecting feature quality. To counter this, we need to further restrict edge creation conditions. Visualization showed clustering tendencies in survey locations; we identify some smaller clusters and calculate their radii, with an estimated radius of about 10 kilometers for each cluster center. Therefore, we stipulate that nodes over 10 kilometers apart would not be connected by an edge. Moreover, we want the edge weight between two nodes to increase as their distance decreased, hence we set the edge weight as the maximum allowable distance (10 kilometers) minus the actual distance between two nodes. Based on these criteria, we add an edge for every pair of nodes that meet the conditions, thus forming the graph.

Next, we need to establish rules for generating a graph feature vector (GFV) for each node through label aggregation from adjacent nodes. We convert each node's neighboring node labels into 0-1 encoded vectors, where the presence of a species is marked as 1 and absence as 0, resulting in each node's label being a 11255-dimensional vector. For any node, the weighted sum of all its neighboring nodes' label vectors and corresponding edge weights constitutes the node's GFV. To avoid the severe imbalance in the numerical distribution of GFV mentioned in the previous section, we adopt a square-root normalization trick similar to that used in attention mechanisms, using the square-root of the degree as the divisor for the GFV. Additionally, to prevent global imbalances in the numerical distribution of GFVs among all nodes, we normalize all nodes' GFVs and reassign them accordingly.

Finally, we add each Survey ID from the test set as nodes to this graph, determining whether to create edges with existing training set nodes based on the established criteria (note that edges should not be generated between test set nodes to avoid affecting the statistics of node degrees). We then aggregate the GFVs for inference on the test set nodes.
The specific calculation formula is as follows:
\FloatBarrier
\begin{equation}
    GFV_i = \frac{\sum_{j=1}^{D_i} L_{ij} \times W_{ij}}{\sqrt{D_i}}
\end{equation}
\begin{equation}
	W_{i j}=10-6731 \times Rad_{i j},
\end{equation}
\FloatBarrier
where $i$ represents the ID of the current node, $j$ represents the ID of the node connected to it, $W$ represents the edge weight between these two nodes, $L$ represents the label vector passed through by the nodes adjacent to the current node, $D$ represents the degree of the current node, and $Rad$ represents the radian distance calculated between two points based on their latitude and longitude.

\FloatBarrier
\begin{algorithm}
\caption{Graph Construction and Feature Vector Aggregation}
\begin{algorithmic}[1]
\State \textbf{Input:} Nodes with attributes, $N$; maximum distance threshold, $d_{\text{max}} = 10$ km.
\State \textbf{Output:} Graph with edges and node feature vectors (GFV).
\Procedure{CreateEdges}{$N$}
    \For{each node $n_i$ in $N$}
        \For{each node $n_j$ in $N$ where $i \neq j$}
            \State Calculate distance, $dist(n_i, n_j)$
            \If{$dist(n_i, n_j) \leq d_{\text{max}}$ and $n_i.\text{year} = n_j.\text{year}$ and $n_i.\text{region} = n_j.\text{region}$}
                \State $weight = d_{\text{max}} - dist(n_i, n_j)$
                \State Add edge $(n_i, n_j)$ with weight $weight$
            \EndIf
        \EndFor
    \EndFor
\EndProcedure

\Procedure{ComputeGFV}{$N$}
    \For{each node $n_i$ in $N$}
        \State Initialize GFV $V_i = 0$
        \State Compute degree $deg(n_i)$ of node $n_i$
        \For{each adjacent node $n_j$}
            \State $V_i \mathrel{{+}{=}} \text{weight}(n_i, n_j) \cdot \text{label}(n_j)$
        \EndFor
        \State Normalize $V_i$ using $\sqrt{deg(n_i)}$
    \EndFor
    \State Normalize all $V_i$ across $N$
\EndProcedure

\State Call \Call{CreateEdges}{$N$}
\State Call \Call{ComputeGFV}{$N$}
\end{algorithmic}
\end{algorithm}
\FloatBarrier

In the second step, we clone the graph with its nodes, edges, weights, years, regions, coordinates, and labels into a new graph, marking training and test set nodes with category labels. We then label the auxiliary nodes from the PO metadata table grouped by Survey ID according to the first step's edge creation conditions and create edges with qualifying test nodes. For each test node, we select the $N$ adjacent training nodes with the highest weights (closest geographical locations) and identify species appearing more than $L$ times, generating a list. We then select the $A$ auxiliary nodes with the highest weights, identify species appearing more than $M$ times, and compile another list. Merging and deduplicating these two lists provides a high-probability species list for each test node, used for correcting model output in post-processing. The default settings are: \( N = 5 \), \( L = 4 \), \( A = 10 \), \( M = 8 \).

\FloatBarrier
\begin{algorithm}
\caption{Complete Graph Construction and Species List Compilation with Sorting}
\begin{algorithmic}[1]
\State \textbf{Input:} Graph $G$ with all attributes
\State \textbf{Output:} Enhanced species lists for test nodes

\Procedure{CloneGraph}{$G$}
    \State $G_{new} \gets$ clone of $G$ with nodes, edges, weights, years, labels
    \State Label nodes in $G_{new}$ as training or test based on category labels
\EndProcedure

\Procedure{LabelAuxiliaryNodes}{$G_{new}$, PO}
    \For{each node in PO grouped by Survey ID}
        \State Apply edge creation conditions from step 1
        \If{node qualifies}
            \State Create edges to test nodes in $G_{new}$
        \EndIf
    \EndFor
\EndProcedure

\Procedure{GenerateSpeciesLists}{$G_{new}$}
    \For{each test node $t$ in $G_{new}$}
        \State Initialize species count maps $Count\_L_t$ and $Count\_M_t$
        
        \State \Comment{Sort and process adjacent training nodes}
        \State Sort all adjacent nodes of $t$ by weight in descending order
        \State $N_t \gets$ top 5 nodes from sorted list
        \For{each node $n$ in $N_t$}
            \For{each species $s$ in $n.labels$}
                \State Increment $Count\_L_t[s]$ by 1
            \EndFor
        \EndFor
        \State $L_t \gets$ species in $Count\_L_t$ where count $> 4$
        
        \State \Comment{Sort and process auxiliary nodes}
        \State Sort all eligible auxiliary nodes by weight in descending order
        \State $A_t \gets$ top 10 nodes from sorted list
        \For{each node $a$ in $A_t$}
            \For{each species $s$ in $a.labels$}
                \State Increment $Count\_M_t[s]$ by 1
            \EndFor
        \EndFor
        \State $M_t \gets$ species in $Count\_M_t$ where count $> 8$
        
        \State \Comment{Merge and deduplicate lists}
        \State $S_t \gets$ merge and remove duplicates from $L_t$ and $M_t$
        
    \EndFor
\EndProcedure

\Procedure{PostProcessOutput}{$G_{new}$}
    \For{each test node $t$ in $G_{new}$}
        \State Use $S_t$ to correct model output
    \EndFor
\EndProcedure

\State Call \Call{CloneGraph}{$G$}
\State Call \Call{LabelAuxiliaryNodes}{$G_{new}$, PO}
\State Call \Call{GenerateSpeciesLists}{$G_{new}$}
\State Call \Call{PostProcessOutput}{$G_{new}$}

\end{algorithmic}
\end{algorithm}
\FloatBarrier

\subsection{Temporal Feature Extraction}
\label{sec3.3}
Another highlight of our work is the development of a temporal feature extraction method based on the Swin-Transformer Block. This approach is inspired by Haixu Wu et al., who used a CNN-based Inception backbone network to extract features from folded time series data in TimesNet\cite{bib9}. Wu and colleagues argue that this method, compared to traditional time series neural networks, is not only better at capturing multi-scale sequential relationships but also has a stronger capability for spatio-temporal information fusion. It shows superior performance across various time series analysis tasks and offers more efficient training and inference. Our goal in processing time series is to obtain higher quality features that are more conducive to modality fusion, rather than making better predictions about future time points. Therefore, we believe that using a visual backbone network to process time series cubes should yield better feature extraction results than traditional time series models.

In the current research on deep learning technology, whether for image processing or time series prediction tasks, methods based on Transformers are considered superior to those based on CNNs. Consequently, we experiment with a backbone network specially designed for extracting temporal features using Swin-Transformer Blocks and Vision Transformer Blocks.

Taking Swin-Transformer as an example, for time series cubes cropped to sizes (4, 18, 12) and (6, 4, 20), we set the Patch size to (3,3) and (2,5), and the Window size to (3,2) and (2,3), respectively. We stack a Swin-Transformer Stage with a depth of 2 and 12 attention heads and another with a depth of 6 and 24 attention heads to create the backbone network for handling our specific time series cubes. The attention function can be defined mathematically by the equation:
\begin{equation}
\text{Attention}(Q, K, V) = \text{SoftMax}\left(\frac{QK^T}{\sqrt{d}} + B\right),
\end{equation}
where $Q$ represents the query matrix, $K$ represents the key matrix,
$V$ represents the value matrix, $d$ is the dimensionality of the keys and queries, typically used for scaling, $B$ represents the positional encoding for time sequences. Unlike the classic Swin-Transformer, where a two-dimensional vector indicates the absolute position of patches in an image, here we employ a one-dimensional vector to represent the position of each patch in a flattened sequence. This modification better adapices to temporal tasks. Apart from that, the meanings of other symbols and formulas are the same as those in the classic Swin-Transformer, and are not reiterated here \cite{bib11}.

We depart from the standard practice of stacking four stages in the Swin-Transformer backbone network because the dimensions of the time series cubes are too small. With each stacked stage, the dimensions for the subsequent stage are halved. Moreover, the Swin-Transformer Block does not accept odd dimensions for input feature maps, making two blocks the most logical configuration for our current time series cube dimensions.

The depths of the two stacked blocks are chosen to be 2 and 6, corresponding to the depths of the second and third stages in the classic Swin-Transformer backbone network. This configuration means that they extract shallow and deep features, respectively. This 1:3 depth ratio has also been adopted by subsequent backbone networks, such as ConvNeXt \cite{bib14}.

We select 12 and 24 as the attention head counts for the two stages, following the counts used in the third and fourth stages of the classic Swin-Transformer backbone network. Increasing the number of attention heads allows the model to independently capture features across more subspaces. Since the feature map sizes entering the third and fourth stages in the original model are already quite small, similar to the size of our time series cubes, having many attention heads does not overly increase the computational burden. Therefore, we use as many attention heads as possible to more comprehensively extract features from the time series cubes. The input and output representations of the final model are as follows:
\begin{equation}
	T = \text{Temporal-Swin-T($U$)},
\end{equation}
where $U\in \mathbb R^{H\times W\times C}$, $H$ represents the number of seasons or months, $W$ represents the number of years, $C$ represents the number of channels, $T$ are the features after Swin-Transformer processing, respectively.
\FloatBarrier
\begin{figure}[h]
	\centering
	\includegraphics[width=1\linewidth]{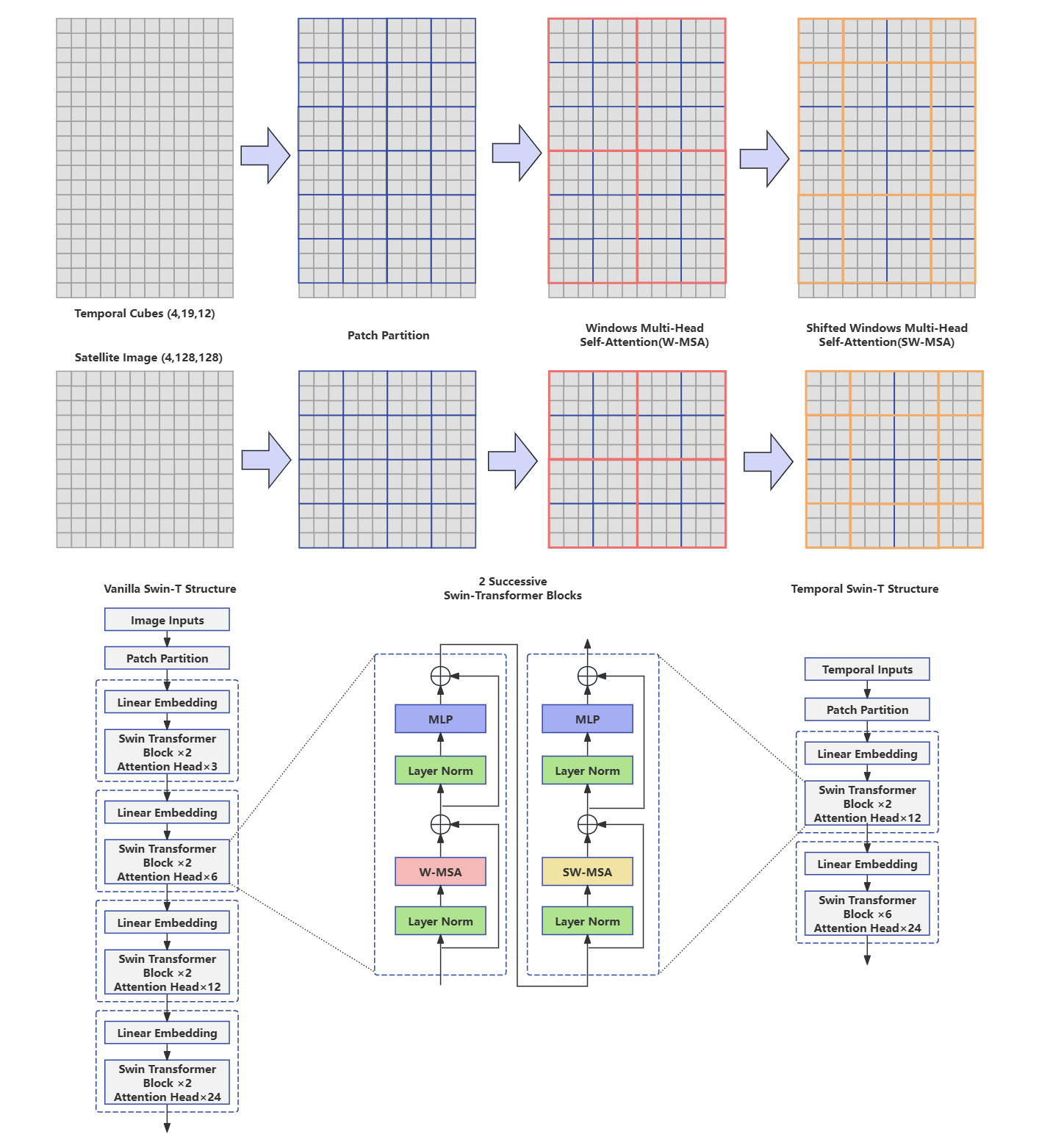}
	\caption{Comparison of Swin-Transformer for temporal cubes processing and image processing}
	\label{fig10}
\end{figure}
\FloatBarrier
We give the comparison diagram to reveal the differences and connections between our model and the classic Swin-Transformer (shown in Figure \ref{fig8}).

Based on this rationale, we also design a Vision Transformer (ViT) backbone network for extracting temporal information.
To validate our approach, we design rigorous comparative experiments using our new temporal feature extraction network to replace the Resnet18 \cite{bib11} used in the baseline for extracting time series features. Given the small size of the time series cubes relative to images, we primarily choose small backbone networks to prevent overfitting and gradient vanishing. 

Our comparative backbone networks include the baseline's ResNet18 \cite{bib10}, the improved version of Inception, Xception41 \cite{bib17} from TimesNet \cite{bib9}, the CNN-based lightweight backbone networks EfficientNet-B0 \cite{bib13} and MoblieNetV3 \cite{bib15}, as well as our designed Swin-Transformer and Vision Transformer backbone networks for time series. The final experimental results demonstrate that our custom-designed temporal feature extraction networks perform optimally, with Swin-Transformer showing the fastest gradient descent and the best results on the private leaderboard.

\subsection{Image Feature Extraction}
\label{sec3.4}
Although this is a multi-task classification problem, the low resolution of satellite images clearly does not allow for accurate prediction of plant species in a given region. Therefore, the primary motivation for processing images is still to extract high-quality features that are conducive to modality fusion. In the baseline, the image feature extraction network employed is the tiny version of Swin-Transformer, which was presented in a Best Paper at ICCV 2021 \cite{bib10}. We note that most models are trained with input sizes of (224,224) or (384,384). To better utilize the weights preserved in the pre-trained model and retain the original information carried by the satellite images, we resize the satellite images from (128,128) to (224,224) before inputting them into the pre-trained model for fine-tuning. Experimental evidence shows that this adjustment significantly enhances the model's performance.

To explore whether there are better alternatives, we conduct comparative experiments with other models such as EfficientNet-B0 \cite{bib13}, ConvNeXt-Base \cite{bib14}, and ViT-Base \cite{bib12}, all pre-trained with an input size of (224,224). The choice of ConvNeXt-Base and ViT-Base is based on their status along with Swin-Transformer as the current state-of-the-art (SOTA) solutions in computer vision backbone networks, and they are comparable to the tiny version of Swin-Transformer in terms of the number of parameters. EfficientNet-B0 was selected partly because it is one of the most powerful feature extraction networks among traditional CNN-based backbones, seen as a superior alternative to the ResNet scheme. Moreover, EfficientNet-B0 is exceptionally lightweight and converges quickly, both in terms of parameter count and training time, allowing for significant optimization of the overall model training and inference time without substantial performance loss.

Our comparative experiments reveal that using the ResNet18 scheme for both temporal and image feature extraction achieved the highest accuracy. However, using Swin-Transformer for temporal feature extraction and EfficientNet-B0 for image feature extraction reduced training time by approximately 75\% without a significant decrease in accuracy, thanks to faster per-epoch training durations and overall faster convergence.
The input image is $I\in \mathbb R^{H\times W\times C}$.
\begin{equation}
	S =\text{Swin-Transformer-Tiny($I$)},
\end{equation}
where $S$ is the feature map of output.$H$ represents the height of satellite images, $W$ represents the width, $C$ represents the number of channels.

\subsection{Hierarchical Cross-Attention Fusion Mechanism(HCAM)}
\label{sec3.5}
Another significant highlight of our work is the introduction of a hierarchical cross-attention fusion mechanism to address the challenge of efficiently fusing feature vectors with varying information densities extracted from different modalities. In the aforementioned steps, we have extracted information ($T$, $S$, $F'$, $G'$) from four modalities, including time-series modal ($U$), satellite imagery modal ($I$), and tabular modal features ($F$), as well as graph modal features ($G$) we constructed and extracted ourselves. $F'$ and $G'$ represent $F$ and $G$ after being processed by fully connected layers.

In terms of information density, satellite imagery modal features are the most dense because the backbone network essentially compresses information carried by multiple channels of an image into a limited set of features for classification mapping via a fully connected layer. Time-series and tabular modal features are less dense, as in our model, we attempt to map them to feature vectors of the same or even higher dimensions. Graph modal features are the sparsest; although we aggregate features from different nodes, the majority of elements in a graph feature vector relative to all 11,255 dimensions are still marked as 0.

\indent Initially, we try to use concatenation for modality fusion similar to the baseline, but due to the high dimensionality and sparsity of the graph feature vectors, the model's loss reduction process is unstable. Consequently, we decide to use cross-attention, more commonly seen in multimodal learning, to attempt fusion of these modalities. However, cross-attention supports only the fusion of two modalities at a time, requiring six uses for pairwise fusion among four modalities, and still necessitating concatenation to integrate each cross-attention output. This not only increases computational overhead but also fails to ensure a controllable reduction in training loss. After multiple adjustments using cross-attention, we determine the optimal modality fusion structure, with specific operations and motivations as follows:

Firstly, the features extracted from the satellite imagery modal are the densest in information. However, observations of the raw data reveal that satellite images primarily provide features of the landscape and vegetation cover, such as color and density of foliage, at the location of the current SurveyID. These features may map to higher-order latent features such as seasonal climate and ecological environment. Meanwhile, the time-series features record climate characteristics and vegetation changes of the area, and the tabular features mainly document the ecological environment characteristics of the area. We consider the time-series and tabular features as two sets of queries, querying keys of climate and vegetation change features, and ecological environment features respectively, both derived from the image features through two parallel linear layers. This setup allows the calculation of attention scores for the time-series and image modalities on climate and vegetation changes, and for the image and tabular modalities on ecological features. The outputs of the cross-attention from these two groups are then concatenated after being calculated from two sets of values mapped from the image features through two parallel linear layers. This concatenated output serves as the final output of the cross-attention calculation for these three modalities. Simultaneously, when the features of the three modalities are fed into this cross-attention module, they are concatenated and mapped to the same dimension as the output of the cross-attention through a linear layer serving as a cutoff, and added together, forming a residual connection. This addition enhances the robustness of the cross-attention module during training. The Cross attention(CA) function can be defined mathematically by the equation:
\begin{gather}
    Q_1 = T \times W_{QT}, \quad Q_2 = F' \times W_{QF}, \notag \\
    K_1 = S \times W_{K1}, \quad K_2 = S \times W_{K2}, \notag \\
    V_1 = S \times W_{V1}, \quad V_2 = S \times W_{V2}, \notag \\
    \begin{aligned}
        A_1 &= \text{Softmax}\left(\frac{Q_1 K_1^T}{\sqrt{d_{k1}}}\right) V_1, \quad 
        A_2 &= \text{Softmax}\left(\frac{Q_2 K_2^T}{\sqrt{d_{k2}}}\right) V_2,
    \end{aligned} \notag \\
    O = \text{Concat}(A_1, A_2) , \quad 
    O_2 = \text{Concat}(T, F, S), \notag \\
    O_{\text{CA}} = O + \text{Linear}(O_2).
\end{gather}
In our initial concept, we consider using the graph modal as the primary modality, given that its features are directly aggregated from the labels of adjacent nodes. Based on the assumptions mentioned at the beginning of section \ref{sec3.2}, these features should closely approximate the current node's label. Therefore, we intend to use the features from other modalities to correct the graph modal, aiming to achieve higher accuracy. However, during training, the difference in information density between the graph modal features and other modal features was too great. Whether through direct concatenation, mapping three modalities' features to the graph feature vector's dimension for addition, or through cross-attention, it was ineffective in guiding the graph feature modal to generate accurate labels for prediction samples (nodes).

In ecology, the composition of vegetation species in a specific spacetime is often determined by factors such as climate conditions, ecological environments, and vegetation changes. Considering that the species appearing in a node's adjacent nodes can represent these three major features of the node's spacetime, albeit in a too sparse representational form, we decide to reduce the dimensionality of the graph feature vector. We compress it to the same dimension as the concatenated vector of the other three modalities, then perform another multi-head cross-attention(MHCA) calculation. Here, the concatenated vector of the other three modal features serves as the Query, with the compressed graph feature vector being mapped as Key and Value by two parallel linear layers, calculating the multi-head cross-attention and outputting. Our rationale for this choice is that the concatenated vector of the three modal features represents the measured characteristics of climate conditions, ecological environment, and vegetation changes in the current spacetime, while the compressed graph feature vector represents the observational results of these characteristics on the plant species combinations we are focusing on. We aim to reveal the causal relationships between them through cross-attention. The Multi-head cross attention function can be defined mathematically by the equation:
\begin{gather}
    G' = \text{Linear}(G) , \quad C = \text{Concat}(F, S, T), \notag \\
    Q = C \times W_Q , \quad K = G' \times W_K, \quad V = G' \times W_V, \notag \\
    O_\text{MHCA}(Q, K, V) = \text{Softmax}\left(\frac{Q K^T}{\sqrt{d_k}}\right) V.
\end{gather}
Ultimately, we concatenated the output of this multi-head cross-attention with the output of the tri-modal cross-attention module mentioned previously, organizing it into the final output through a fully connected layer. 
\begin{gather}
    O_{\text{final}} = \text{Linear}(\text{Concat}(O_{\text{MHCA}}, O_{\text{CA}})) ,\notag \\
    O_{\text{output}} = \text{Sigmoid}(O_{\text{final}}).
\end{gather}
This approach significantly improves the model's performance during training. Observing the loss curves for the training and validation sets during the training process, it is evident that the model's overfitting was markedly reduced, and the loss on the validation set could decrease further.

Throughout the process, we use two cross-attention layers, performing three cross-attention calculations. Initially, we select the image modal feature with the highest information density and calculate cross-attention with the time-series and tabular modal features, which have relatively high information densities. Subsequently, we calculate the cross-attention between the features of the three modalities and the compressed graph modal feature in one go. The selection of modal features for each cross-attention layer reflects a hierarchical relationship, thus we name this approach the Hierarchical Cross-Attention Fusion Mechanism. The schematic diagram of HCAM is as follows:
\FloatBarrier
\begin{figure}[h]
	\centering
	\includegraphics[width=1\linewidth]{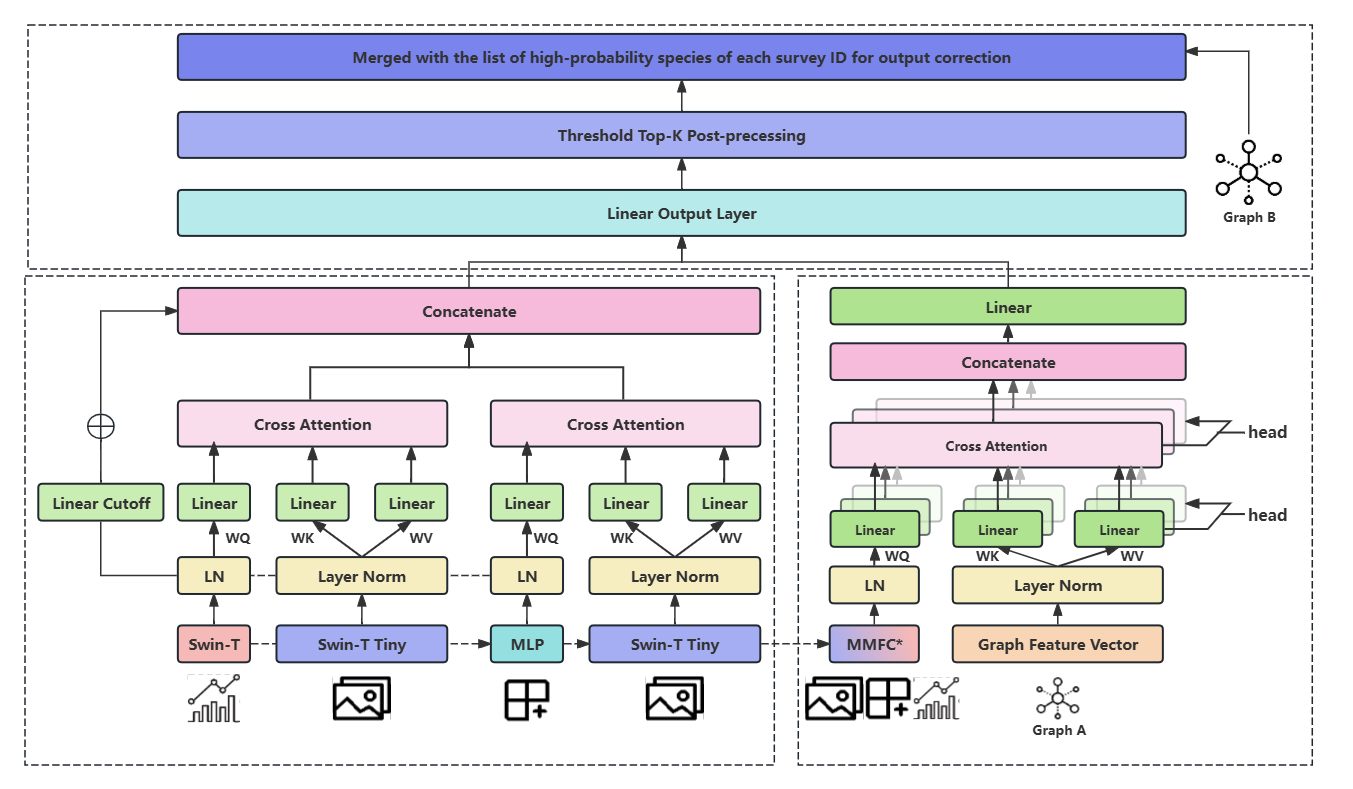}
	\caption{Schematic of the structure of the proposal model}
	\label{fig8}
\end{figure}
\FloatBarrier
\subsection{Mix up +10 Fold Cross Fusion training strategy}
\label{sec3.6}

We adopt the Mix up training strategy provided in the baseline, which is a very common method for data augmentation during training. Specifically, this method involves randomly shuffling the order of training data and labels in the current batch and then performing a weighted addition with the unshuffled training data and labels. 
The input matrices \( U \), \( I \), \( F \), and \( G \), as well as the label matrix \( L \), are shuffled to create \( \tilde{T} \), \( \tilde{S} \), \( \tilde{F} \), \( \tilde{G} \), and \( \tilde{L} \):
\begin{equation}
\begin{aligned}
T_{\text{mix}} &= \lambda T + (1 - \lambda) \tilde{T}, \\
S_{\text{mix}} &= \lambda S + (1 - \lambda) \tilde{S}, \\
F_{\text{mix}} &= \lambda F + (1 - \lambda) \tilde{F}, \\
G_{\text{mix}} &= \lambda G + (1 - \lambda) \tilde{G}, \\
L_{\text{mix}} &= \lambda L + (1 - \lambda) \tilde{L}.
\end{aligned}
\end{equation}

This approach enhances the robustness of the training process, improves the generalization performance of the model, smooths out the distribution of samples across different categories, and makes originally sparse labels relatively dense.

Moreover, to fully utilize the training data and reduce model overfitting, we employ a ten-fold cross-fusion technique to train the model. The concept of ten-fold cross-fusion is an improvement over ten-fold cross-validation. The specific operation involves dividing the dataset into ten parts, with each part serving once as the validation set, while the other nine parts are used to train a brand new model. The logits output by these ten new models are then averaged. However, this approach also results in a training efficiency about ten times lower than before. To address this drawback, I reason that the training datasets used for each model are highly overlapping, and except for the first model, the training of the subsequent nine models can be considered as fine-tuning the first model using a slightly altered dataset. Motivated by this, we optimize our training strategy. For the first model, we initialize parameters and train it using an early stopping strategy. For each subsequent model, we clone the parameters of the first model and fine-tune it on a newly combined training set. This method reduces the number of training epochs for subsequent models, thereby enhancing training efficiency. { The original dataset is $D$; we divide it into ten parts $\{D_1,D_2,\ldots,D_{10}\}$. 
We train a model $M_i$ by setting each $D_i$ as the test dataset: $M_i=\mathrm{Train}\left(D\backslash D_i\right)$}.
Then we compute the average of logits from these models:
\begin{equation}
	L = \frac{1}{10} \sum_{i=1}^{10} M_i(x),
\end{equation}
where $x$ is the model input. The training process and results visualization for each model are as follows:

\FloatBarrier
\begin{figure}[h]
	\centering
	\includegraphics[width=1\linewidth]{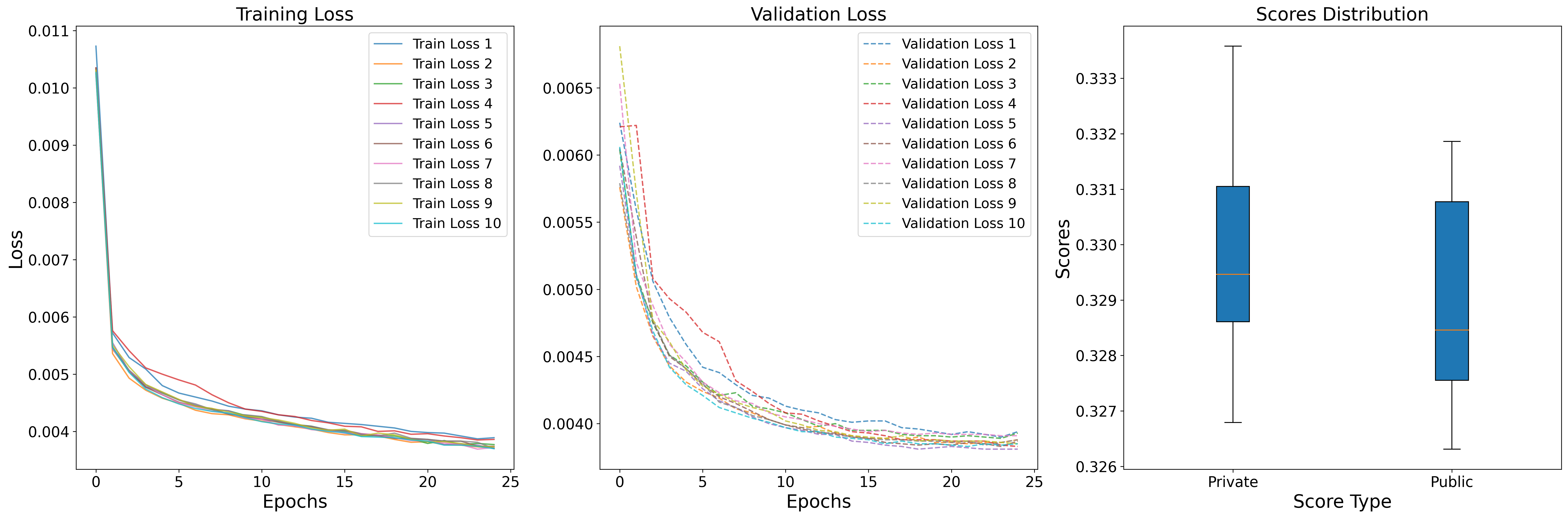}
	\caption{Training loss and validation loss plots of 10 fold training}
	\label{fig11}
\end{figure}
\FloatBarrier

The binary cross-entropy (BCE) loss is one of the most common loss functions for multi-label learning \cite{bib8}. For an observed data point $(x_n, y_n)$, including full positive and negative classes, the BCE loss is calculated as follows:

\begin{equation}
    L_{\text{BCE}}(f_n, y_n) = -\frac{1}{L} \sum_{i=1}^L \left[ \mathbf{1}_{y_{n,i}=1} \log(f_{n,i}) + \mathbf{1}_{y_{n,i}=0} \log(1 - f_{n,i}) \right],
\end{equation}
where $f_n = f(x_n) \in [0, 1]$ is the model predicted probability of presence for species $i$ under input $x_n$, and $\mathbf{1}$ denotes the indicator function, i.e., $\mathbf{1}_k = 1$ if assertion $k$ is true, or 0 otherwise.

\subsection{Post-Processing: Threshold Top-K and Output Correction}
\label{sec3.7}
Based on the official requirements, the final formula for calculating the score is as follows:
\begin{equation}
	F_1=\frac{1}{N}\sum_{i=1}^{N}\frac{{\mathrm{TP}}_i}{{\mathrm{TP}}_i+\left({\mathrm{FP}}_i+{\mathrm{FN}}_i\right)/2},
\end{equation}
\begin{equation}
	\text {where }\left\{\begin{array}{c}
		\mathrm{TP}_{i}=\text { Number-of-predicted-species-truly-present, i.e., }\left|\widehat{Y}_{i} \cap Y_{i}\right| \\
		\mathrm{FP}_{\mathrm{i}}=\text { Number-of-species predicted-but-absent, i.e.. }\left|\widehat{Y}_{1} \backslash \mathrm{Y}_{\mathrm{i}}\right| \\
		\mathrm{FN}_{\mathrm{i}}=\text { Number-of-species not-predicted-but-present, i.e., }\left|\mathrm{Y}_{\mathrm{i}} \backslash \widehat{Y}_{1}\right|.
	\end{array}\right.
\end{equation}

In the baseline, the classic multi-class task method of Top-K is used to filter the model's output. Typically, in conventional Top-K, a value for K is manually set or initially a range is predefined, within which K is enumerated to observe which K yields the best performance on the validation set inference results, and this K is then applied to the test set inference process.  

However, we observe drawbacks with this method. Some Survey IDs contain dozens or even hundreds of species. Although these species might have high probabilities in the model's output, they are truncated if they do not rank within the top K. Additionally, some test set Survey IDs, even with low probabilities for each species, are still forced to output the top K species by probability rank.

Therefore, we improve the Top-K algorithm by setting a range of thresholds (0.1 to 0.5, in steps of 0.01) for each possible K to filter out species with predicted probabilities below these thresholds. By exhaustively combining K and threshold values and comparing the scores of the validation set outputs processed through them, we can identify the optimal pair of K and threshold.

{Let the probability of the model output be $P=\{p_1,p_2,\ldots,p_{11255}\}$, where $p_i$ is the predicted probability of species $i$. We filter the output by setting the threshold $\theta$ and K values:}
{\begin{equation}
		S = \{i |p_i\ > \theta\}\cup\text{Top-K}(P),
\end{equation}}
where $S$ is the set of filtered species.

\begin{figure}[h]      
	\centering
	\setlength{\abovecaptionskip}{0pt}    
	\subfigure[Heatmap]{  
			\includegraphics[width=0.45\linewidth]{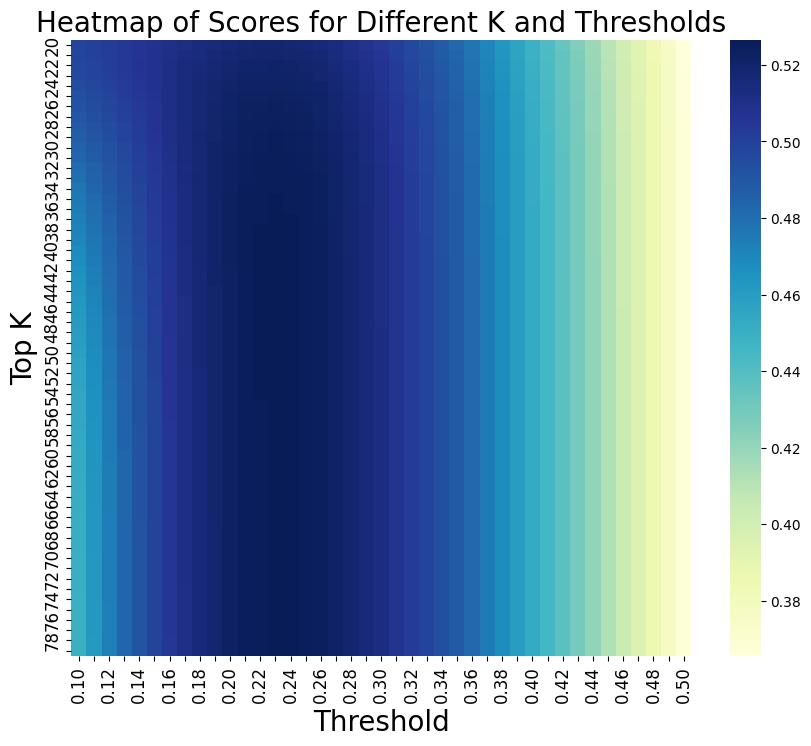}
			\label{fig91} 	}
	\subfigure[3D Plot]{
			\includegraphics[width=0.5\linewidth]{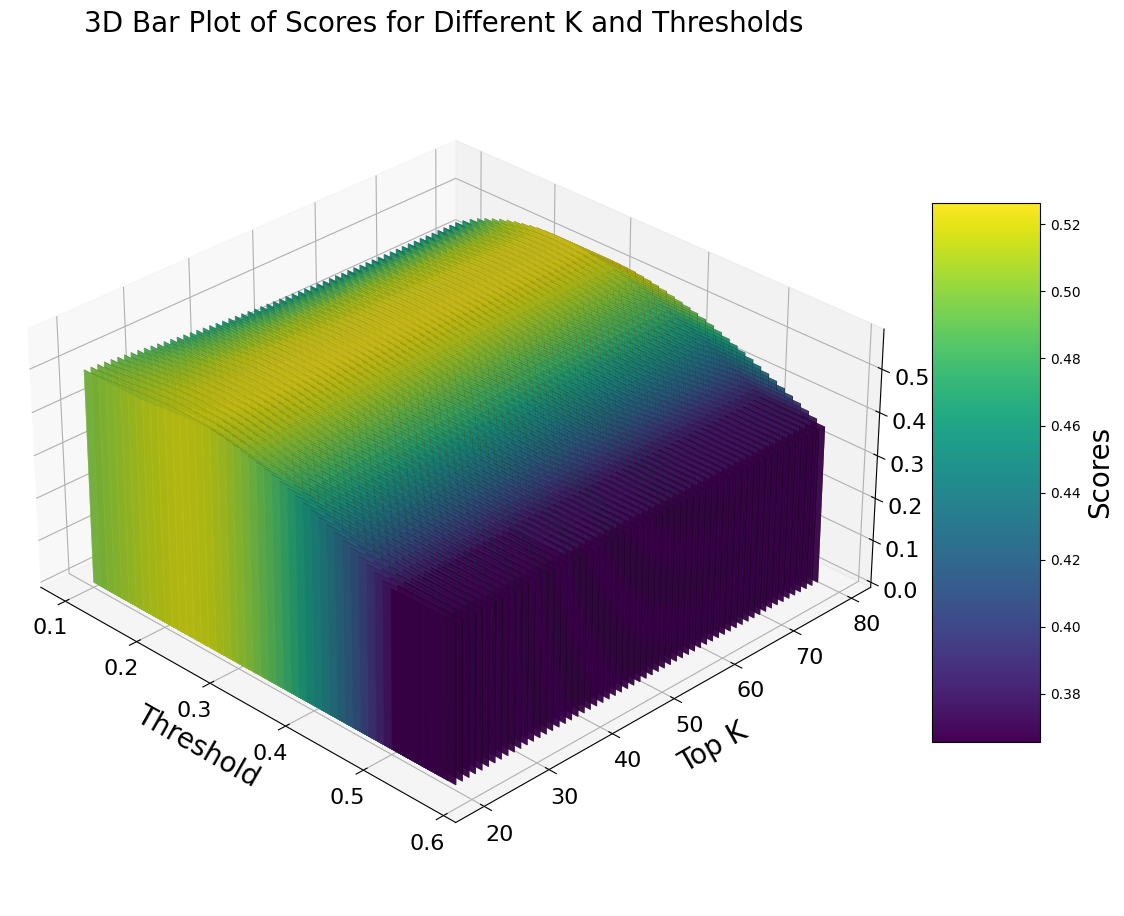}
			\label{fig92}  	}    
	\caption{Heatmap of the threshold and K values correspond to the validation set score}
	\label{fig9}   
\end{figure}

 We can see that the darker colored regions represent higher scores for this pair of parameters. In the end, we choose the optimal K of 44 and the optimal threshold of 0.23 (shown in Figure\ref{fig9}).
\FloatBarrier
\FloatBarrier

We note that there are 11,255 species IDs, but only 5,016 appear in PA, with the remainder in PO, prompting us to correct the model's data through mining of the PO data. The specific approach involves merging and deduplicating the list of high-probability species for each test set node obtained in section \ref{sec3.2} with the model's prediction results to produce the final output. In addition, We find that generating a high-probability species list using only the PO data (auxiliary nodes) is more consistently beneficial compared to generating a high-probability species list using both PO and PA data for result correction.

\section{Experiments}
\label{sec4}
In this chapter, we present the experiments conducted to validate the superiority of the Proposal model and analyze the experimental results. We omit the hyperparameter tuning process in the paper because each backbone network corresponds to different model hyperparameters and training hyperparameters. Given the limited time, it is challenging to prove the optimality of the selected hyperparameters through grid search. Instead, we judge the current parameters' potential to cause overfitting or underfitting by observing the gradient descent process, or whether the current parameter combination results in lower loss and higher scores compared to previous combinations.We perform exploratory tuning for each model involved in the comparative and ablation experiments to ensure that the current hyperparameter combination is the best among all attempts. We also employ an early stopping strategy, which minimizes the impact of hyperparameter changes on training when there is no significant overfitting or underfitting. 

At the end of this chapter, we provide a table of the hyperparameters used for the final submission.We select the following metrics to analyze the model's performance, including public and private leaderboard scores on the official test set. Additionally, since we ultimately need to fuse the logits output by the models, we introduce the loss on the validation and training sets. Since the calculation of BCE essentially equals the sum of entropy and KL divergence, it describes the difference between the probability distribution of the model output and the distribution of the true labels. Therefore, the logits of models with lower validation and training set losses result in better fusion effects than models with lower validation scores but higher losses. Finally, we explore how to effectively lightweight the model, using the training time per epoch as a metric.

Our experiments are all conducted on the Colab platform running on an A100 instance. The specific computational resources include 83.5GB of memory and 40GB of GPU memory.

To facilitate the customization of model parameters, we use the Timm library instead of the torchvision library to instantiate models. Surprisingly, the baseline reconstructed based on the Timm library achieved a better private leaderboard score compared to the official baseline (the official baseline's private leaderboard score is 0.31626, while the baseline improved by optimizing the Top-K strategy achieved a private leaderboard score of 0.32359).
\subsection{Comparative Experiments}
\label{sec4.1}
For the satellite image feature extraction network, the comparative experiments are as follows:

\FloatBarrier
\begin{table}[h]
	\caption{Comparative experiment for satellite image feature extraction network}
	\label{tab1}
    \begin{tabular}{ccccccc}
    \toprule
    Module & Val score & Public score & Private score & Val loss & Training loss & Time cost \\
    \midrule
    Swin-T-Tiny &{\color{blue}0.42110} & 0.32774 & 0.32816 &{\color{blue}0.00379} & 0.00353 &{\color{blue}271}\\
    \textbf{ViT-Base} & \textbf{0.42645} & \textbf{0.33376} & \textbf{0.33280} & \textbf{0.00379} & 0.00349 & 550\\
    ConvNeXt-Base &  0.42066 & 0.33104 & 0.32792 & 0.00381  & 0.00361 &705\\
    {\color{blue}EfficientNet-B0} &0.41114  & {\color{blue}0.33316} & {\color{blue}0.33085} & 0.00393 & 0.00338 & \textbf{130} \\
    \bottomrule
\end{tabular}
\end{table}
\FloatBarrier

Bold  indicates the best score for this metric, while blue, if present, indicates a score close to the best. Training loss is used to help determine if overfitting has occurred and does not represent model performance. We observe that, based on the scores, the best-performing model is ViT-Base, followed by EfficientNet-B0. However, our goal is to identify the models most suitable for logits fusion, so we also focus on the loss. We find that ViT-Base and Swin-T-Tiny have very similar validation losses, making ViT-Base the preferred model.
In terms of training time, EfficientNet-B0 and Swin-T-Tiny are the most efficient models. Considering the number of epochs to convergence, EfficientNet-B0 reaches the overfitting threshold in almost half the time of Swin-T-Tiny, but its test set loss is higher than that of Swin-T-Tiny. Therefore, EfficientNet-B0 can be considered a successful attempt at model lightweighting. However, for the highest score after fusion, the focus should still be on ViT-Base and Swin-T-Tiny.

Because the comparative experiments of the temporal feature extraction network and the image feature extraction backbone network are conducted simultaneously, the corresponding image feature extraction network during the comparative experiments of the temporal feature extraction network is still the Swin-T-Tiny from the baseline. The comparative experiments are as follows:

\FloatBarrier
\begin{table}[h]
	\caption{Comparative experiment for time series feature extraction network}
	\label{tab2}
	\begin{tabular}{ccccccc}
		\toprule
		Module & Val score & Public score & Private score & Val loss & Training loss & Time cost \\
		\midrule
		ResNet18 & 0.42110 & 0.32774 & 0.32816 & 0.00379 & 0.00353 & \textbf{271} \\
		\textbf{Swin-T} & 0.42319 & \textbf{0.33428} & \textbf{0.33476} & \textbf{0.00377} & 0.00362 & 279 \\
		ViT & 0.41794 & 0.31680 & 0.31772 & 0.00390 & 0.00339 & 277 \\
		EfficientNet-B0 & 0.42083 & 0.31940 & 0.31996 & 0.00381 & 0.00344 & 292 \\
		MobileNet-V3 & 0.41479 & 0.31435 & 0.31649 & 0.00390 & 0.00339 & 284 \\
        \textbf{Xception41} & \textbf{0.42737} & 0.32488 & 0.32216 & \textbf{0.00377} & 0.00331 & 302 \\
		\bottomrule
	\end{tabular}
\end{table}
\FloatBarrier

Based on the experimental results, the Swin-T backbone network we proposed for extracting temporal features achieves the highest private leaderboard score and the lowest validation loss. It is also the only backbone network that scored higher than the Baseline. In terms of training time and convergence epochs, it is comparable to the Baseline. Therefore, we believe that the Swin-T backbone network can replace ResNet18 in the Baseline.
\subsection{Ablation Studies}
\label{sec4.2}
Based on the conclusions drawn from the comparative experiments, we initially replace the backbone networks for image and temporal feature extraction starting from the Baseline model. Subsequently, we attempt to use the Swin-T backbone network for temporal feature extraction and ViT-Base for satellite image feature extraction, but encounter difficulties with gradient descent. We analyze this issue and found that the significant difference in the number of parameters between the two networks, along with the misalignment in the gradient descent speeds, led to this problem. Since we cannot effectively resolve this issue, we opt to use Swin-T-Tiny, which has a similar validation loss, as the satellite image feature extraction network. We also try replacing Swin-T-Tiny with EfficientNet-B0, but this only accelerates training efficiency without improving the scores. Therefore, we determine that the optimal backbone network for temporal feature extraction is Swin-T, and for satellite image feature extraction is Swin-T-Tiny.

Next, we attempt to introduce the Graph modality. We find that models incorporating the Graph modality showed significant improvement in validation scores and the lowest validation loss so far, but the public and private leaderboard scores decrease. Upon examining the output, we discover that including the Graph modality makes the model more aware of some minority classes that are often overlooked, significantly increasing their logits values. However, under the Top-K output rules, these minority classes still do not rank high enough, and not all boosted minority classes are present. Those that do not appear show up in the logits values of some Survey IDs with generally low confidence. This result in better validation loss but lower private leaderboard scores. This phenomenon inspire us to propose the threshold Top-K as an improved post-processing algorithm.

We then try using HCAM (Hierarchical Cross-Attention Mechanism) for feature fusion. We find that the validation loss of models fused with HCAM slightly increased, but the public and private leaderboard scores return to the levels before integrating the Graph modality. This demonstrates that HCAM effectively optimizes the logits distribution while improving model scores.

To verify whether the model incorporating GFV (Graph Feature Vector) + HCAM is indeed more suitable for ten-fold cross-validation fusion, we conduct ten-fold cross-validation training on the model with GFV + HCAM and the baseline model using Swin-T for temporal feature extraction and Swin-Transformer-Tiny for image feature extraction with feature concatenation for modality fusion. We find that the model incorporating GFV + HCAM scored higher on the leaderboard and validation set compared to the control group.

It is important to note that the validation set mentioned in the ten-fold cross-validation method is the same as the validation set used in previous ablation experiments. However, since the complete training set is used in ten-fold cross-validation, the data in the validation set is actually included in the training set. Therefore, the validation set scores here are higher than in experiments without ten-fold cross-validation, and should only be compared between experiments using the same ten-fold cross-validation method.

In the official competition submission, we overlook checking the output of the model incorporating GFV + HCAM and simply judge the failure of our GFV construction and HCAM design based on the public leaderboard score. Consequently, we choose the previous model with ten-fold cross-validation training and applied post-processing. However, when completing this paper, we realize that the model incorporating GFV + HCAM might have more potential for fusion. Subsequent experiments confirme that this approach can indeed provide further improvements in leaderboard scores.

We acknowledge that the current results lack persuasive power for the ablation experiments on HCAM and Graph modality features. Besides the reasons we analyzed, HCAM and the MLP used for compressing Graph modality features are the parts of our model structure that are more sensitive to parameter changes. Our lack of time for sufficient hyperparameter analysis is also one reason for the unsatisfactory ablation experiment data. In future work, we will explore the optimal parameters and structures for these two sub-networks.

Finally, we apply our proposed post-processing methods to the model outputs. From the ablation experiments, it is evident that both Threshold Top-K and corrections based on PO data improved model scores. However, since the validation set cannot contain species present only in PO, the validation set scores are inaccurate in this context. Furthermore, the post-processing operations do not affect the model's training and inference efficiency, so they do not need to be included in the time cost comparison.

The final results are summarized in the table below(\noindent * denotes that the expected improvement is not achieved, thus this approach is abandoned):

\FloatBarrier
\begin{table}[h]
	\caption{Ablation studies results}
	\label{tab3}
\begin{tabular}{ccccc}
	\toprule
	Module/Tricks & Val score & Public score & Private score & Time cost \\
	\midrule
	Baseline & 0.42110 & 0.32774 & 0.32816 & 271 \\
	Swin-T replace ResNet & 0.42319  & 0.33428 & 0.33476 & 279 \\
	{\color{blue}EfficientNet replace Swin-Transformer-Tiny*}  &0.41114  & 0.33316 & 0.33085 &  130\\
    EfficientNet and Swin-T* & 0.40824 & 0.30872 & 0.31349 & 137 \\
    Graph Modal Feature Vector (GFV) & 0.43950 & 0.32251 &0.32543  &  290\\
	HCAM with GFV &  0.42489 & 0.33186 & 0.33358 & 292 \\
	10 Fold Cross Fusion without HCAM and GFV* & 0.49120 & 0.34790 & 0.34625 &  -\\
    10 Fold Cross Fusion with HCAM and GFV & 0.50455 & 0.35170 & 0.34994 &  -\\
	Threshold Top-K &  0.52655 & 0.36428 & 0.36121 & - \\
    \textbf{Output Correction (Final Model)} & - & \textbf{0.36478} & \textbf{0.36242} & - \\
	\bottomrule
\end{tabular}
\end{table}
\FloatBarrier

To provide a more intuitive confirmation of our analysis on the training process losses of different models discussed in this chapter, we visualize the training processes of all models. This facilitates the comparison of the number of epochs required for different models to reach their optimal loss (shown in Figure\ref{fig9}).
\FloatBarrier
\begin{figure}[h]
	\centering
	\includegraphics[width=1\linewidth]{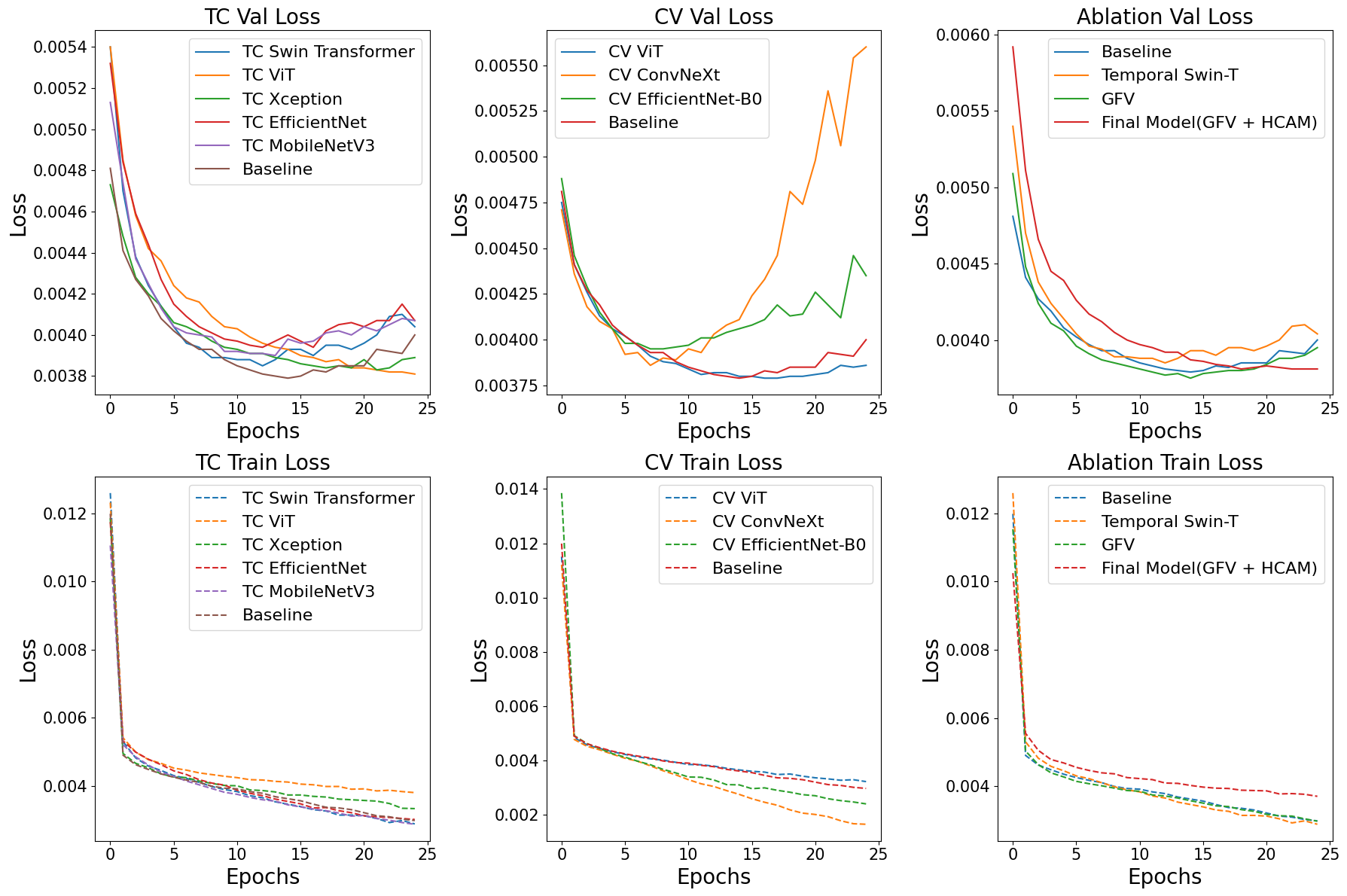}
	\caption{Training loss and validation loss plots of Ablation and comparative experiments}
	\label{fig12}
\end{figure}
\FloatBarrier

Where 'CV' stands for the comparative experiments of the visual modality, while 'TC' represents the comparative experiments of the Temporal Cubes modality.

\section{Conclusion and Discussion}
\label{sec5}
\subsection{Conclusion}
\label{sec5.1}
Our comprehensive comparative experiments demonstrate that we select the most suitable feature extraction backbone networks for each modality's data. Through rigorous ablation studies, we prove that each improvement we proposed incrementally enhance the model's performance, ultimately achieving a score of 0.36242 on the private leaderboard and in third place(The leaderboard showed 0.35292, but we continue to optimize some parameters while doing the experiment to finally get 0.36242). In addition to proposing high-scoring solutions, we also introduce a lightweight model and an efficient training strategy. Without significantly sacrificing accuracy, we significantly reduce the training time for the ten-fold cross-fusion model (by more than 50\%) and the time for a single epoch and the total number of epochs (by approximately 75\%). Thus, we manage to train ten models for logits fusion in roughly the same amount of time it previously takes to train one baseline model, achieving prediction results far exceeding the baseline score. It is worth mentioning that we average the outputs of all models that score higher than the baseline and apply post-processing, ultimately achieving a private leaderboard score of 0.36501.

In summary, from a theoretical perspective, our research provides new insights into the extraction of temporal information for modality fusion tasks, namely folding it into a 2D matrix and extracting features through a visual backbone network. We also design a robust fusion method for features extracted from multiple modalities with varying information densities, using a hierarchical cross-attention mechanism to dilute features from high to low density progressively. Additionally, we propose a graph-based feature construction method and output correction post-processing algorithm for the multi-task classification task of species prediction under specific spacetime, which often involves extremely unbalanced or sparse labels. From an application perspective, our research is of significant importance in fields such as ecology, agriculture, environmental protection, and climate change studies.
\subsection{Future Work}
\label{sec5.2}
Despite of our comprehensive comparative and ablation experiments, our work may be further explored in the following way:

\begin{itemize}
\item We will further utilize the organized PO data, noting that there are some high-quality publishers within PO whose surveys of the same ID often contain many species. Additionally, the competition provides supplementary data about other modalities for Survey IDs in PO. We believe that these Survey IDs appearing in PO can be selected through certain logical criteria to serve as training samples for weakly supervised learning, incorporated into PA. This approach will better leverage crowdsourced data, reduce the workload of data collection, and enhance model accuracy.
    \item Referring to past programs, highly ranked teams would extract the rasters around the Survey ID geographic location from the tiff files provided by EnvironmentalRasters as images to be processed. However, due to arithmetic and time constraints, we finally give up on implementing this scheme, and we plan to use this part of the data in our subsequent work to realize a leap in model performance.
	\item Our current model establishes graph relationships solely for feature aggregation and result calibration. In future work, we plan to use graph neural networks to replace the current method of manually setting weights for adjacent nodes in feature aggregation. This will support the use of weakly supervised and semi-supervised learning training strategies to progressively correct labels of weakly supervised and semi-supervised nodes, thereby improving training outcomes.
	\item We hope to introduce NAS technology to optimize the hyperparameters of our 2D time series feature extraction network based on Swin-Transformer and the hierarchical cross-attention mechanism, further enhancing the model's performance.
\end{itemize}
\newpage
\begin{acknowledgments}
The data for this paper is organized and published by INRIA. We express our gratitude to all the institutions and individuals involved in data collection and processing, including but not limited to the Global Biodiversity Information Facility (GBIF, www.gbif.org), NASA, Soilgrids, and the Ecodatacube platform. Additionally, this project has received funding from the European Union's Horizon Research and Innovation program under grant agreements No. 101060639 (MAMBO project) and No. 101060693 (GUARDEN project) \cite{geolifeclef2024,lifeclef2024}.
All authors contribute helpful ideas during the course of the competition and participate in writing and revising the paper, so all authors are co-first authors. All of the authors, as corresponding authors, are obliged to reply to emails to provide readers with the relevant code and data of this work and explain the details of the work. Among them, Haixu Liu, as the first corresponding author, is responsible for the necessary communication for the publication of the article.
Our final submission CSV download link is as follows: \href{https://drive.google.com/file/d/12oUBGB5ENy16lIkyQpvsDaH9nQe1NpI8/view?usp=sharing}{submission\_036242.csv} and 
\href{https://drive.google.com/file/d/1XzPQQ2bLucriDTaGzcPyjwdRNKsUeGQd/view?usp=sharing}{submission\_036501.csv}. The link to the code we use to run and obtain the final submission is as follows: \href{https://colab.research.google.com/drive/1-Zo4J6RyKm0a-cPQLmlP4gv9Fqqglw-B?usp=sharing}{code}.

\end{acknowledgments}

\bibliography{touche24-paper-literature.bib}

\begin{thebibliography}{20}
\expandafter\ifx\csname natexlab\endcsname\relax\def\natexlab#1{#1}\fi
\providecommand{\url}[1]{\texttt{#1}}
\providecommand{\href}[2]{#2}
\providecommand{\path}[1]{#1}
\providecommand{\DOIprefix}{doi:}
\providecommand{\ArXivprefix}{arXiv:}
\providecommand{\URLprefix}{URL: }
\providecommand{\Pubmedprefix}{pmid:}
\providecommand{\doi}[1]{\href{http://dx.doi.org/#1}{\path{#1}}}
\providecommand{\Pubmed}[1]{\href{pmid:#1}{\path{#1}}}
\providecommand{\bibinfo}[2]{#2}
\ifx\xfnm\relax \def\xfnm[#1]{\unskip,\space#1}\fi
\bibitem[{Picek et~al.(2024)Picek, Botella, Servajean, Deneu, Marcos~Gonzalez, Palard, Larcher, Leblanc, Estopinan, Bonnet, and Joly}]{geolifeclef2024}
\bibinfo{author}{L.~Picek}, \bibinfo{author}{C.~Botella}, \bibinfo{author}{M.~Servajean}, \bibinfo{author}{B.~Deneu}, \bibinfo{author}{D.~Marcos~Gonzalez}, \bibinfo{author}{R.~Palard}, \bibinfo{author}{T.~Larcher}, \bibinfo{author}{C.~Leblanc}, \bibinfo{author}{J.~Estopinan}, \bibinfo{author}{P.~Bonnet}, \bibinfo{author}{A.~Joly},
\newblock \bibinfo{title}{Overview of {GeoLifeCLEF} 2024: Species presence prediction based on occurrence data and high-resolution remote sensing images},
\newblock in: \bibinfo{booktitle}{Working Notes of CLEF 2024 - Conference and Labs of the Evaluation Forum}, \bibinfo{year}{2024}.
\bibitem[{Joly et~al.(2024)Joly, Picek, Kahl, Go{\"e}au, Espitalier, Botella, Deneu, Marcos, Estopinan, Leblanc, Larcher, \v{S}ulc, Hr\'{u}z, Servajean et~al.}]{lifeclef2024}
\bibinfo{author}{A.~Joly}, \bibinfo{author}{L.~Picek}, \bibinfo{author}{S.~Kahl}, \bibinfo{author}{H.~Go{\"e}au}, \bibinfo{author}{V.~Espitalier}, \bibinfo{author}{C.~Botella}, \bibinfo{author}{B.~Deneu}, \bibinfo{author}{D.~Marcos}, \bibinfo{author}{J.~Estopinan}, \bibinfo{author}{C.~Leblanc}, \bibinfo{author}{T.~Larcher}, \bibinfo{author}{M.~\v{S}ulc}, \bibinfo{author}{M.~Hr\'{u}z}, \bibinfo{author}{M.~Servajean}, et~al.,
\newblock \bibinfo{title}{Overview of {LifeCLEF} 2024: Challenges on species distribution prediction and identification},
\newblock in: \bibinfo{booktitle}{International Conference of the Cross-Language Evaluation Forum for European Languages}, \bibinfo{organization}{Springer}, \bibinfo{year}{2024}.
\bibitem[{Leblanc et~al.(2022)Leblanc, Joly, Lorieul, Servajean, and Bonnet}]{bib6}
\bibinfo{author}{C.~Leblanc}, \bibinfo{author}{A.~Joly}, \bibinfo{author}{T.~Lorieul}, \bibinfo{author}{M.~Servajean}, \bibinfo{author}{P.~Bonnet},
\newblock \bibinfo{title}{Species distribution modeling based on aerial images and environmental features with convolutional neural networks},
\newblock in: \bibinfo{booktitle}{CLEF (Working Notes)}, \bibinfo{year}{2022}, pp. \bibinfo{pages}{2123--2150}.
\bibitem[{He et~al.(2016)He, Zhang, Ren, and Sun}]{bib10}
\bibinfo{author}{K.~He}, \bibinfo{author}{X.~Zhang}, \bibinfo{author}{S.~Ren}, \bibinfo{author}{J.~Sun},
\newblock \bibinfo{title}{Deep residual learning for image recognition},
\newblock in: \bibinfo{booktitle}{Proceedings of the IEEE conference on computer vision and pattern recognition}, \bibinfo{year}{2016}, pp. \bibinfo{pages}{770--778}.
\bibitem[{Kellenberger and Tuia(2022)}]{bib7}
\bibinfo{author}{B.~Kellenberger}, \bibinfo{author}{D.~Tuia},
\newblock \bibinfo{title}{Block label swap for species distribution modelling},
\newblock in: \bibinfo{booktitle}{CLEF (Working Notes)}, \bibinfo{year}{2022}, pp. \bibinfo{pages}{2103--2114}.
\bibitem[{Ung et~al.(2023)Ung, Kojima, and Wada}]{bib8}
\bibinfo{author}{H.~Q. Ung}, \bibinfo{author}{R.~Kojima}, \bibinfo{author}{S.~Wada},
\newblock \bibinfo{title}{Leverage samples with single positive labels to train cnn-based models for multi-label plant species prediction},
\newblock \bibinfo{journal}{Working Notes of CLEF}  (\bibinfo{year}{2023}).
\bibitem[{Joly et~al.(2023)Joly, Botella, Picek, Kahl, Go{\"e}au, Deneu, Marcos, Estopinan, Leblanc, Larcher et~al.}]{lifeclef2023}
\bibinfo{author}{A.~Joly}, \bibinfo{author}{C.~Botella}, \bibinfo{author}{L.~Picek}, \bibinfo{author}{S.~Kahl}, \bibinfo{author}{H.~Go{\"e}au}, \bibinfo{author}{B.~Deneu}, \bibinfo{author}{D.~Marcos}, \bibinfo{author}{J.~Estopinan}, \bibinfo{author}{C.~Leblanc}, \bibinfo{author}{T.~Larcher}, et~al.,
\newblock \bibinfo{title}{Overview of lifeclef 2023: evaluation of ai models for the identification and prediction of birds, plants, snakes and fungi},
\newblock in: \bibinfo{booktitle}{International Conference of the Cross-Language Evaluation Forum for European Languages}, \bibinfo{organization}{Springer}, \bibinfo{year}{2023}, pp. \bibinfo{pages}{416--439}.
\bibitem[{Potapov et~al.(2020)Potapov, Hansen, Kommareddy, Kommareddy, Turubanova, Pickens, and Ying}]{bib1}
\bibinfo{author}{P.~Potapov}, \bibinfo{author}{M.~C. Hansen}, \bibinfo{author}{I.~Kommareddy}, \bibinfo{author}{A.~Kommareddy}, \bibinfo{author}{S.~Turubanova}, \bibinfo{author}{A.~Pickens}, \bibinfo{author}{Q.~Ying},
\newblock \bibinfo{title}{Landsat analysis ready data for global land cover and land cover change mapping},
\newblock \bibinfo{journal}{Remote Sensing} \bibinfo{volume}{12} (\bibinfo{year}{2020}) \bibinfo{pages}{426}.
\bibitem[{Witjes et~al.(2022)Witjes, Parente, van Diemen, Hengl, Landa, Brodsk{\'y}, and Glu{\v{s}}ica}]{bib2}
\bibinfo{author}{M.~Witjes}, \bibinfo{author}{L.~Parente}, \bibinfo{author}{C.~J. van Diemen}, \bibinfo{author}{T.~Hengl}, \bibinfo{author}{M.~Landa}, \bibinfo{author}{L.~Brodsk{\'y}}, \bibinfo{author}{L.~Glu{\v{s}}ica},
\newblock \bibinfo{title}{A spatiotemporal ensemble machine learning framework for generating land use/land cover time-series maps for europe (2000--2019) based on lucas, corine and glad landsat},
\newblock \bibinfo{journal}{PeerJ} \bibinfo{volume}{10} (\bibinfo{year}{2022}) \bibinfo{pages}{e13573}.
\bibitem[{Witjes et~al.(2023)Witjes, Parente, Kri{\v{z}}an, Hengl, and Antoni{\'c}}]{bib3}
\bibinfo{author}{M.~Witjes}, \bibinfo{author}{L.~Parente}, \bibinfo{author}{J.~Kri{\v{z}}an}, \bibinfo{author}{T.~Hengl}, \bibinfo{author}{L.~Antoni{\'c}},
\newblock \bibinfo{title}{Ecodatacube.eu: analysis-ready open environmental data cube for europe},
\newblock \bibinfo{journal}{PeerJ} \bibinfo{volume}{11} (\bibinfo{year}{2023}) \bibinfo{pages}{e15478}.
\bibitem[{Karger et~al.(2017)Karger, Conrad, B{\"o}hner, Kawohl, Kreft, Soria-Auza, and Kessler}]{bib4}
\bibinfo{author}{D.~N. Karger}, \bibinfo{author}{O.~Conrad}, \bibinfo{author}{J.~B{\"o}hner}, \bibinfo{author}{T.~Kawohl}, \bibinfo{author}{H.~Kreft}, \bibinfo{author}{R.~W. Soria-Auza}, \bibinfo{author}{M.~Kessler},
\newblock \bibinfo{title}{Climatologies at high resolution for the earth’s land surface areas},
\newblock \bibinfo{journal}{Scientific data} \bibinfo{volume}{4} (\bibinfo{year}{2017}) \bibinfo{pages}{1--20}.
\bibitem[{Karger et~al.(2018)Karger, Conrad, B{\"o}hner, Kawohl, Kreft, Soria-Auza, and Kessler}]{bib5}
\bibinfo{author}{D.~N. Karger}, \bibinfo{author}{O.~Conrad}, \bibinfo{author}{J.~B{\"o}hner}, \bibinfo{author}{T.~Kawohl}, \bibinfo{author}{H.~Kreft}, \bibinfo{author}{R.~W. Soria-Auza}, \bibinfo{author}{M.~Kessler},
\newblock \bibinfo{title}{Data from: Climatologies at high resolution for the earth’s land surface areas},
\newblock \bibinfo{journal}{EnviDat}  (\bibinfo{year}{2018}).
\bibitem[{Vaswani et~al.(2017)Vaswani, Shazeer, Parmar, Uszkoreit, Jones, Gomez, Kaiser, and Polosukhin}]{bib18}
\bibinfo{author}{A.~Vaswani}, \bibinfo{author}{N.~Shazeer}, \bibinfo{author}{N.~Parmar}, \bibinfo{author}{J.~Uszkoreit}, \bibinfo{author}{L.~Jones}, \bibinfo{author}{A.~N. Gomez}, \bibinfo{author}{{\L}.~Kaiser}, \bibinfo{author}{I.~Polosukhin},
\newblock \bibinfo{title}{Attention is all you need},
\newblock \bibinfo{journal}{Advances in neural information processing systems} \bibinfo{volume}{30} (\bibinfo{year}{2017}).
\bibitem[{Wu et~al.(2022)Wu, Hu, Liu, Zhou, Wang, and Long}]{bib9}
\bibinfo{author}{H.~Wu}, \bibinfo{author}{T.~Hu}, \bibinfo{author}{Y.~Liu}, \bibinfo{author}{H.~Zhou}, \bibinfo{author}{J.~Wang}, \bibinfo{author}{M.~Long},
\newblock \bibinfo{title}{Timesnet: Temporal 2d-variation modeling for general time series analysis},
\newblock in: \bibinfo{booktitle}{The eleventh international conference on learning representations}, \bibinfo{year}{2022}.
\bibitem[{Liu et~al.(2021)Liu, Lin, Cao, Hu, Wei, Zhang, and Guo}]{bib11}
\bibinfo{author}{Z.~Liu}, \bibinfo{author}{Y.~Lin}, \bibinfo{author}{Y.~Cao}, \bibinfo{author}{H.~Hu}, \bibinfo{author}{Y.~Wei}, \bibinfo{author}{Z.~Zhang}, \bibinfo{author}{B.~Guo},
\newblock \bibinfo{title}{Swin transformer: Hierarchical vision transformer using shifted windows},
\newblock in: \bibinfo{booktitle}{Proceedings of the IEEE/CVF international conference on computer vision}, \bibinfo{year}{2021}, pp. \bibinfo{pages}{10012--10022}.
\bibitem[{Liu et~al.(2022)Liu, Mao, Wu, Feichtenhofer, Darrell, and Xie}]{bib14}
\bibinfo{author}{Z.~Liu}, \bibinfo{author}{H.~Mao}, \bibinfo{author}{C.~Y. Wu}, \bibinfo{author}{C.~Feichtenhofer}, \bibinfo{author}{T.~Darrell}, \bibinfo{author}{S.~Xie},
\newblock \bibinfo{title}{A convnet for the 2020s},
\newblock in: \bibinfo{booktitle}{Proceedings of the IEEE/CVF conference on computer vision and pattern recognition}, \bibinfo{year}{2022}, pp. \bibinfo{pages}{11976--11986}.
\bibitem[{Chollet(2017)}]{bib17}
\bibinfo{author}{F.~Chollet},
\newblock \bibinfo{title}{Xception: Deep learning with depthwise separable convolutions},
\newblock in: \bibinfo{booktitle}{Proceedings of the IEEE conference on computer vision and pattern recognition}, \bibinfo{year}{2017}, pp. \bibinfo{pages}{1251--1258}.
\bibitem[{Tan and Le(2019)}]{bib13}
\bibinfo{author}{M.~Tan}, \bibinfo{author}{Q.~Le},
\newblock \bibinfo{title}{Efficientnet: Rethinking model scaling for convolutional neural networks},
\newblock in: \bibinfo{booktitle}{International conference on machine learning}, \bibinfo{organization}{PMLR}, \bibinfo{year}{2019}, pp. \bibinfo{pages}{6105--6114}.
\bibitem[{Howard et~al.(2017)Howard, Zhu, Chen, Kalenichenko, Wang, Weyand, and Adam}]{bib15}
\bibinfo{author}{A.~G. Howard}, \bibinfo{author}{M.~Zhu}, \bibinfo{author}{B.~Chen}, \bibinfo{author}{D.~Kalenichenko}, \bibinfo{author}{W.~Wang}, \bibinfo{author}{T.~Weyand}, \bibinfo{author}{H.~Adam},
\newblock \bibinfo{title}{Mobilenets: Efficient convolutional neural networks for mobile vision applications},
\newblock \bibinfo{journal}{arXiv preprint arXiv:1704.04861}  (\bibinfo{year}{2017}).
\bibitem[{Dosovitskiy et~al.(2020)Dosovitskiy, Beyer, Kolesnikov, Weissenborn, Zhai, Unterthiner, and Houlsby}]{bib12}
\bibinfo{author}{A.~Dosovitskiy}, \bibinfo{author}{L.~Beyer}, \bibinfo{author}{A.~Kolesnikov}, \bibinfo{author}{D.~Weissenborn}, \bibinfo{author}{X.~Zhai}, \bibinfo{author}{T.~Unterthiner}, \bibinfo{author}{N.~Houlsby},
\newblock \bibinfo{title}{An image is worth 16x16 words: Transformers for image recognition at scale},
\newblock \bibinfo{journal}{arXiv preprint arXiv:2010.11929}  (\bibinfo{year}{2020}).

\end{thebibliography}


\end{document}